\pdfoutput=1
\documentclass[11pt]{article}

\usepackage[]{acl}

\usepackage{times}
\usepackage{latexsym}
\usepackage{graphicx}
\usepackage{bbold}
\usepackage{tikz}
\usepackage{multirow}
\usepackage{caption}
\usepackage{subcaption}
\usepackage{amsmath}
\usepackage{amssymb}
\usepackage{xurl}

\usepackage{comment}
\usepackage{float}
\usepackage{chngpage}
\usepackage{cuted}
\usepackage{booktabs}
\usepackage{algorithm}
\usepackage{algpseudocode}
\usepackage{tabulary}
\usepackage{tabularray}
\usepackage{array}
\usepackage{xcolor}

\def\testnext#1#2{#1\ifx-#2\color{purple}\else\ifcat#20\color{teal}\else\fi\fi#2}

\usepackage[normalem]{ulem}
\useunder{\uline}{\ul}{}

\usepackage[T1]{fontenc}
\usepackage[utf8]{inputenc}
\usepackage{microtype}

\title{Comparing Explanation Faithfulness between Multilingual and Monolingual Fine-tuned Language Models}

\author{Zhixue Zhao \quad  Nikolaos Aletras \\
        Department of Computer Science, University of Sheffield \\
        United Kingdom \\
\texttt{\{zhixue.zhao, n.aletras\}@sheffield.ac.uk}}

\begin{document}

\maketitle

\begin{abstract}

In many real natural language processing application scenarios, practitioners not only aim to maximize predictive performance but also seek faithful explanations for the model predictions. Rationales and importance distribution given by feature attribution methods (FAs) provide insights into how different parts of the input contribute to a prediction. Previous studies have explored how different factors affect faithfulness, mainly in the context of monolingual English models. On the other hand, the differences in FA faithfulness between multilingual and monolingual models have yet to be explored. Our extensive experiments, covering five languages and five popular FAs, show that FA faithfulness varies between multilingual and monolingual models. We find that the larger the multilingual model, the less faithful the FAs are compared to its counterpart monolingual models.
Our further analysis shows that the faithfulness disparity is potentially driven by the differences between model tokenizers.\footnote{Our code is available \url{https://github.com/casszhao/multilingual-faith}.}

\end{abstract}

\section{Introduction}

Feature attribution methods (FAs) are commonly used for ranking input tokens according to their importance to a model's prediction \citep{kindermans2016investigating, sundararajan2017axiomatic, deyoung-etal-2020-eraser}. Subsequently, the top-k ranked tokens are selected to form a rationale.
The faithfulness of a FA method refers to what extent its token importance scores and selected rationales actually reflect the model's inner reasoning mechanism~\cite{jacovi-goldberg-2020-towards}.

\begin{figure}[h]
\centering
    \includegraphics[trim={0 .5cm 0 1cm}, width=\columnwidth]{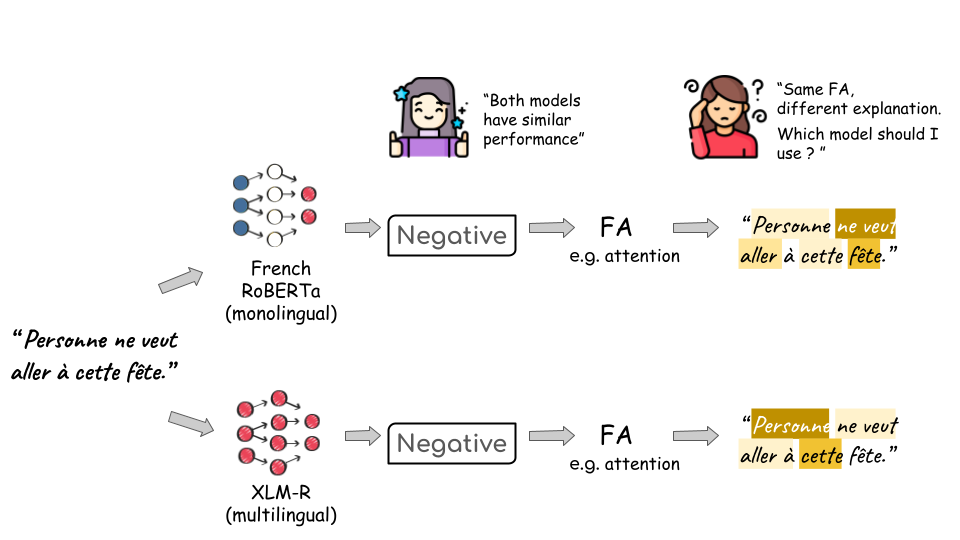}
    \caption{Model explanations given by the same feature attribution method, e.g. attention, for multilingual (XLM-R) and monolingual (French RoBERTa) models for the same task (sentiment analysis in FR).  
}
    \label{fig:teaser}
\end{figure}

Previous work has mainly studied faithfulness in the context of monolingual models, i.e. especially English \citep{atanasova-etal-2020-diagnostic, bastings-filippova-2020-elephant,chan-etal-2022-comparative}. Specifically, monolingual studies have investigated the impact of out-of-domain data~\cite{chrysostomou-aletras-2022-empirical}, adversarial attacks \citep{sinha-etal-2021-perturbing, zhao-etal-2022-really} and temporal shifts~\cite{zhao-etal-2022-impact} on the faithfulness of FAs.   
On the other hand, existing studies on interpreting multilingual models' behavior and their representations \citep{rama-etal-2020-probing,serikov-etal-2022-universal,gonen-etal-2022-analyzing} have not investigated the faithfulness of FAs.

As shown in Figure \ref{fig:teaser}, even for the same input (``Personne ne veut aller à cette fête.'', i.e. ``Nobody wants to go to this party.'' in English), model prediction and FA, the token importance scores can be substantially different between multi- and monolingual models. 
This indicates that the models follow different inner processes for making predictions. It is unclear whether this difference is generally shared among input examples or even across other languages and models. Given that the performance of multilingual models might be on par with monolingual counterparts in various languages \citep{rust-etal-2021-good, su-etal-2022-rocbert}, this leaves practitioners in a dilemma between choosing multilingual or monolingual models when the application scenario requires extracting faithful explanations for the model predictions. 

In this paper, we seek to answer \textit{if there is a faithfulness disparity of FAs when applied to multi- and monolingual models.} Our main contributions are as follows:

\begin{itemize}
    \item We perform a large empirical study across tasks in five languages, five popular FAs and two groups of monolingual and multilingual models;
    \item Our results reveal that the degree of faithfulness disparity can be attributed to the size of the models, i.e. FAs tend to give less faithful rationales for larger multilingual models, compared to their monolingual counterparts;
    \item Our analysis further shows that the discrepancies in faithfulness are potentially driven by differences in tokenization rather than how these models semantically process the input. For example, the more aggressive tokenization results in a larger faithfulness discrepancy between mono- and multilingual models. 
\end{itemize}

\section{Related Work}

\subsection{Faithfulness of monolingual models}
Faithfulness measures if a rationale extracted with a given FA, accurately reflects the model's internal reasoning process \citep{ribeiro-etal-2016-trust, deyoung-etal-2020-eraser, jacovi-goldberg-2020-towards,pezeshkpour-etal-2021-empirical}.\footnote{Plausibility evaluates the extent to which the rationale aligns with human understanding \citep{jacovi-goldberg-2020-towards} and it is out of the scope of our study.}

On the one hand, existing faithfulness studies on monolingual models mainly focus on English. \citet{sinha-etal-2021-perturbing} and \citet{zhao-etal-2022-really} explored how adversarial attacks affect the faithfulness of FAs by swapping tokens to create new inputs with the same semantics. \citet{bastings-etal-2022-will} introduced ground truth, i.e. fully faithful rationales, with specific but meaningless tokens, to evaluate faithfulness.
\citet{chrysostomou-aletras-2022-empirical} investigated the impact of out-of-domain data on faithfulness, while \citet{zhao-etal-2022-impact} studied the effect of temporal concept drift on faithfulness. On the other hand, an increasing number of multilingual language models are made available for different languages 
\citep{antoun2020arabert, chan-etal-2020-germans, canete2020spanish, le-etal-2020-flaubert-unsupervised}, but there is no empirical evidence that a FA is equally faithful between monolingual models and their counterpart multilingual models.

\subsection{Interpretability of multilingual models}

Previous studies on multilingual models focus on probing or analyzing their hidden representations, which are not directly related to the faithfulness of model explanations. \citet{santy-etal-2021-bertologicomix} monitored the changes of attention heads in multilingual models when the model is further fine-tuned on monolingual and bilingual corpora.
\citet{rama-etal-2020-probing} probed the representations of mBERT (multilingual BERT) between languages and they found that their distances correlate most with phylogenetic and geographical distances between languages.
\citet{gonen-etal-2022-analyzing} analyzed the gender representations of multilingual models. 
\citet{rust-etal-2021-good} studied the difference of multilingual models in processing different languages. They found that languages adequately represented in the multilingual model's vocabulary exhibit negligible performance decreases over their monolingual counterparts.  
\citet{morger-etal-2022-cross} examined the correlation between human focus and model relative word importance on monolingual and multilingual language models. 

Rather than studying the faithfulness of multilingual models, \citet{zaman-belinkov-2022-multilingual} proposed a faithfulness evaluation method for multilingual models. It assumes that an interpretation system is unfaithful if it provides different interpretations for similar inputs and outputs where the similar inputs have the same meaning in different languages without comparing mono- and multilingual models.

\subsection{Performance comparison of monolingual and multilingual models}

Previous work has compared the performance of monolingual and multilingual language models focusing on mBERT and BERT variants~\citep{ronnqvist-etal-2019-multilingual,nozza2020mask,vulic-etal-2020-probing,rust-etal-2021-good}.  
\citet{vulic-etal-2020-probing} specifically investigated how lexical knowledge extraction strategies impact performance between mono- and multilingual models, while
\citet{rust-etal-2021-good}  further investigated the impact of tokenizers. A general observation drawn from these studies is that when the mono- and multilingual models have similar architectures and training objectives, their predictive performance is comparable regardless of the difficulty of the task. Multilingual models' performance is often considered to suffer from the \textit{``curse of multilinguality''} \citep{conneau-etal-2020-unsupervised, pfeiffer-etal-2022-lifting}, i.e. the phenomenon of overall performance decrease on monolingual as well as cross-lingual tasks beyond a certain number of languages. To the best of our knowledge, no study has investigated how the curse of multilinguality impacts FAs' faithfulness.

\section{Experiments}

Our aim is to compare FA faithfulness between mono- and multilingual models across tasks and languages. For this purpose, we experiment with models of similar architectures and pre-training objectives following \citet{rust-etal-2021-good}. The main differences between mono- and multilingual models are the tokenizers, supported vocabularies and the pre-training corpora. Using this setting allows for a realistic comparison between models, given the fact that in a real world scenario, a practitioner would choose between off-the-shelf, already pre-trained mono- or multilingual models without considering any specific implementation details (e.g. pre-training data).
We compare models in various downstream tasks across a spectrum of typologically diverse and widely spoken languages, i.e. English, Chinese, Spanish, French and Hindi.

\renewcommand*{\arraystretch}{1.0}
\begin{table}[!t]
\small
\centering
\resizebox{\columnwidth}{!}{%
\begin{tabular}{llp{2.3cm}lll}
\hline
\textbf{Language} & \textbf{Model} & \textbf{Pre-training Corpus} & \textbf{\#Tokens} & \textbf{Vocab} & \textbf{Params} \\ \hline

 & mBERT & Wiki-100 & 3.3B & 106K & 167M \\
 \multirow{-2}{*}{\textbf{Multi}} & XLM-R & CC-100 & 167B & 250K & 278M \\ 
\hline
\hline
 & BERT & Wikipedia, BookCorpus & 3.3B & 30K & 109M \\
\multirow{-0.5}{*}{\textbf{English (EN)}} & RoBERTa & BookCorpus, CC-News, OpenWebText, Stories & 40B & 50K & 125M\\ 
\hline

 & BERT & Wikipedia & 0.4B & 21K & 103M \\
\multirow{-2}{*}{\textbf{Chinese (ZH)}} & RoBERTa & Wikipedia & 0.4B & 21K & 102M \\ \hline

 & BERT & Wikipedia, OPUS & 3B & 31K & 110M \\
\multirow{-3}{*}{\textbf{Spanish (ES)}} & RoBERTa & Web crawl & 	135B & 50K & 125M \\ \hline

 & BERT & Europeana & 11B & 32K & 111M \\
\multirow{-1}{*}{\textbf{French (FR)}} & RoBERTa & Wikipedia, CC-100 & 59B  & 50K & 124M \\ \hline

 & BERT & L3Cube & 0.3B & 52K & 126M \\ 
\multirow{-1}{*}{\textbf{Hindi (HI)}} & RoBERTa & mC4, OSCAR, IndicNLP &  1.5B  & 52K & 83M \\ \hline
\end{tabular}%
}
\caption{Overview of models across languages.}
\label{table:models}
\end{table}

\subsection{Models}

\paragraph{Multilingual models.}

We use two popular multilingual models: (1) \textbf{mBERT}, a multilingual version of BERT~\citep{devlin-etal-2019-bert} trained on text from 104 languages in Wikipedia; and (2) \textbf{XLM-R} \citep{conneau-etal-2020-unsupervised}, a multilingual version of RoBERTa~\citep{liu2019roberta} trained on text from 100 languages in the Common Crawl corpus. 

\paragraph{Monolingual models.}

For each language, we include its corresponding \textbf{monolingual BERT} and \textbf{RoBERTa} models respectively. Table \ref{table:models} provides an overview of all models across languages. 

\subsection{Datasets}

Due to the lack of identical datasets in multiple languages, we include a variety of tasks that are similar. We experiment with (1) sentiment analysis; (2) topic classification;
(3) reading comprehension; (4) paraphrase identification; and natural language inference. Table \ref{tab:datasets} summarizes datasets used in this paper. \footnote{Following \citet{su-etal-2022-rocbert}, we use the small version of ChnSentiCorp data. Following \citet{le-etal-2020-flaubert-unsupervised}, we sample 2,000 examples from the original French CSL dataset as the training set and another 2,400 examples for development and testing. We repeat the same for Hindi CSL and Spanish CSL. Further, for tasks without a published test and development sets, we split the original dataset using a 80:10:10 ratio for train, development and test with the same label distribution.}

\label{app:datasets}

\begin{table*}[!h]
\resizebox{\textwidth}{!}{%
\begin{tabular}{@{}llllllll@{}}
\toprule
\textbf{Language} & \textbf{Language Family} & \textbf{Dataset} & \textbf{Task} & \textbf{Training set size} & \textbf{Avg length} & \textbf{Metrics} & \textbf{Papers}\\ \midrule
% English
 &  & SST & Sentiment analysis & 6,920 / 872 / 1,821 & 17 & F1 & \citet{socher-etal-2013-recursive}\\
 &  & Agnews & Topic classification & 102,000 / 18,000 / 7,600 & 36 & F1  & \citet{del2005ranking}\\
\multirow{-3}{*}{\textbf{English}} & \multirow{-3}{*}{Indo-European} & MultiRC & Reading Comprehension & 24,029 / 3,214 / 4,848 & 290/17 & F1  & \citet{deyoung-etal-2020-eraser,jain-etal-2020-learning}\\ \midrule
% Chinese
 &  & Ant & Reading Comprehension & 30,018 / 4,316 / 4,316 & 13/13 & Accuracy & \citet{su-etal-2022-rocbert}\\
 &  & KR & Keyword Recognition & 17,000 / 3,000 / 3,000 & 266/29 & Accuracy  & \citet{su-etal-2022-rocbert}\\
\multirow{-3}{*}{\textbf{Chinese}} & \multirow{-3}{*}{Sino-Tibetan} & ChnSentiCorp & Sentiment analysis & 2,000 / 1,200 / 1,200 & 107 & Accuracy  & \citet{su-etal-2022-rocbert}\\ \midrule
% Spanish
 &  & CSL & Sentiment analysis & 2,000 / 1,200 / 1,200 & 27 & Accuracy & \citet{keung-etal-2020-multilingual}\\
 &  & PAWS-X & Paraphrase Identification & 49,400 / 2,000 / 2,000 & 20/20 & Accuracy & \citet{yang-etal-2019-paws}\\
\multirow{-3}{*}{\textbf{Spanish}} & \multirow{-3}{*}{Indo-European} & XNLI & Natural Language Inference & 393,000 / 5,010 / 2,490 & 19/9 & Accuracy & \citet{conneau-etal-2020-unsupervised}\\ \midrule
% French
 &  & CSL & Sentiment analysis & 2,000 / 1,200 / 1,200 & 28 & Accuracy & \citet{le-etal-2020-flaubert-unsupervised,keung-etal-2020-multilingual}\\
 &  & PAWS-X & Paraphrase Identification & 49,400 / 2,000 / 2,000 & 20/20 & Accuracy & \citet{yang-etal-2019-paws},\citet{le-etal-2020-flaubert-unsupervised},\citet{canete-etal-2022-albeto}\\
\multirow{-3}{*}{\textbf{French}} & \multirow{-3}{*}{Indo-European} & XNLI & Natural Language Inference & 393,000 / 5,010 / 2,490 & 20/10 & Accuracy & \citet{le-etal-2020-flaubert-unsupervised}, \citet{conneau-etal-2020-unsupervised},\citet{canete-etal-2022-albeto}\\ \midrule
% Hindi
 &  & BBC NLI & Natural Language Inference & 15,552 / 2,580 / 2,592 & 7/5 & Accuracy & \citet{uppal-etal-2020-two}\\
 &  & News Topic & Topic classification & 15,552 / 2,580 / 2,592 & 13 & F1 & \citet{uppal-etal-2020-two}\\
\multirow{-3}{*}{\textbf{Hindi}} & 
\multirow{-3}{*}{Indo-Aryan} & XNLI & Natural Language Inference & 392,702 / 2,490 / 5,010 & 21/10 & Accuracy & \citet{conneau-etal-2020-unsupervised}\\ \midrule

\end{tabular}%
}
\caption{Datasets summary. For tasks requiring two inputs, e.g. paraphrase identification and language inference tasks, the average text lengths are shown separately for the first and second input as \textit{length 1 / length 2}.}
\label{tab:datasets}
\end{table*}

\subsection{Implementation details}
We fine-tune each model using the hyperparameters from the original papers describing the corresponding models and tasks. If these are not available, we use a batch size of 16 and a learning rate of 1e-5 with an early stopping over five epochs. Full implementation details are given in Appendix \ref{app:model_hyperparameters}.

\subsection{Feature attribution methods}\label{sec:FAs}

We experiment with five popular FAs since it has been shown that there is no single best FA across models and tasks~\cite{atanasova-etal-2020-diagnostic}:\footnote{Our aim is not to exhaustively benchmark various FAs but to explore their faithfulness between mono- and multilingual models across different languages and tasks.}

\begin{itemize}
\item{\bf Attention ($\alpha$):} Importance is computed using the corresponding normalized attention score of the CLS token from the last layer \citep{jain-etal-2020-learning}.

\item{\bf Scaled attention ($\alpha\nabla\alpha$):} Similar to $\alpha$, but the attention score is scaled by its corresponding gradient \citep{serrano-smith-2019-attention}.

\item{\bf InputXGrad ($x\nabla x$):} It attributes importance by multiplying the input with its gradient computed with respect to the predicted class \citep{kindermans2016investigating, atanasova-etal-2020-diagnostic}.

\item{\bf Integrated Gradients (IG):} This FA ranks input tokens by computing the integral of the gradients taken along a straight path from a baseline input (i.e. zero embedding vector) to the original input \citep{sundararajan2017axiomatic}.

\item{\bf DeepLift (DL):} It computes token importance according to the difference between the activation of each neuron and a reference activation, i.e. a zero embedding vector \citep{pmlr-v70-shrikumar17a}.
\end{itemize}

We also include a baseline that randomly assigns importance scores to each token ({\bf Random}).

\subsection{Faithfulness evaluation}

\paragraph{Hard Sufficiency \& Comprehensiveness.} 
Sufficiency (Suff) and comprehensiveness (Comp) are two commonly used metrics for evaluating faithfulness~\citep{deyoung-etal-2020-eraser} using hard input perturbation.

Suff aims to capture the difference in predictive likelihood between retaining only the rationale $p(\hat{y}|\mathcal{R})$ and the full text model $p(\hat{y}| \mathbf{X})$. We use the normalized version for a fairer comparison across models \citep{carton-etal-2020-evaluating}:
\begin{equation}
\label{equ:Norm_Suff}
\small
\begin{aligned}
    \text{Suff}(\mathbf{X}, \hat{y}, \mathcal{R}) = 1 - max(0, p(\hat{y}| \mathbf{X})- p (\hat{y}|\mathcal{R})) \\
    \text{Normalized Suff}(\mathbf{X}, \hat{y}, \mathcal{R}) = \frac{\text{Suff}(\mathbf{X}, \hat{y}, \mathcal{R}) - \text{Suff}(\mathbf{X}, \hat{y}, 0)}{1 - \text{Suff}(\mathbf{X}, \hat{y}, 0)}
 \end{aligned}
\end{equation}

\noindent where $\text{S}(\mathbf{x}, \hat{y}, 0)$ is the sufficiency of a baseline input (zeroed out sequence) and $\hat{y}$ is the model predicted class using the full text $\mathbf{x}$ as input.

Comp assesses how much information the rationale holds by measuring changes in predictive likelihoods when removing the rationale $p(\hat{y}|\mathbf{X}_{\backslash\mathcal{R}})$. The normalized version is defined as: 
\begin{equation}
\label{equ:Norm_Comp}
    \small
    \begin{aligned}
        \text{Comp}(\mathbf{X}, \hat{y}, \mathcal{R}) = max(0, p(\hat{y}| \mathbf{X})- p (\hat{y}|\mathbf{X}_{\backslash\mathcal{R}})) \\\\
        \text{Normalized Comp}(\mathbf{X}, \hat{y}, \mathcal{R}) = \frac{\text{Comp}(\mathbf{X}, \hat{y}, \mathcal{R})}{1 - \text{Suff}(\mathbf{X}, \hat{y}, 0)}
    \end{aligned}
\end{equation}

Following \citet{deyoung-etal-2020-eraser}, we use the Area Over the Perturbation Curve (AOPC) for normalized Suff and Comp across different rationale lengths (10\%, 20\%, and 50\%) by taking the average, similar to \citet{deyoung-etal-2020-eraser} and \citet{chan-etal-2022-comparative}.\footnote{For tasks of average token length over 200, we evaluate rationale ratios of 1\%, 5\%, and 10\% instead, to keep the rationales relatively short.}

\paragraph{Soft Sufficiency \& Comprehensiveness.} Soft sufficiency (\textbf{Soft-Suff}) and comprehensiveness (\textbf{Soft-Comp}) use a soft input perturbation criterion to measure faithfulness \citep{zhao-aletras-2023-incorporating}. Each token is perturbed proportionally to its importance score assigned by a FA instead of being fully retained or removed. The `soft' version of these metrics has been found to be more robust compared to their `hard' counterparts. 

The final scores for the four metrics are computed after being divided by their corresponding random baseline. Therefore, values greater than one denote higher than random faithfulness (the higher, the more faithful).

%%%%%%%%%%%%%%%%%%%%

\section{Results}\label{sec:results_shown_in_the_begining}

Our experiments include two multilingual and ten monolingual models, five FAs, and 15 tasks. Specifically, we test four models (two multilingual and two monolingual) on three tasks, using five FAs in each language. This results in 480 faithfulness evaluation cases for each language, 2400 cases for five languages in total. We report accuracy and F1 of all models in Appendix \ref{app:full_results}. 

\begin{table*}[!t]
\centering
\scriptsize
%\resizebox{0.8\textwidth}{!}{%
\begin{tabular}{lc|c|cc|cc|c|cc|cc}
\toprule
 & \multirow{2}{*}{} & \multicolumn{5}{c|}{BERT \& mBERT} & \multicolumn{5}{c}{RoBERTa \& XLM-R} \\ \cmidrule(l){3-12} 
Lang & Model & Acc & Suff & Comp & \multicolumn{1}{c}{S-Suff} & \multicolumn{1}{c|}{S-Comp} & Acc & Suff & Comp & \multicolumn{1}{c}{S-Suff} & \multicolumn{1}{c}{S-Comp} \\ \midrule

\multirow{2}{*}{EN} & Mono & \textbf{0.847} & 1.146 & 1.525 & 1.172 & 1.201 & \textbf{0.852} & \textbf{1.306} & \textbf{1.588} & 1.207 & \textbf{1.200} \\
 & Multi & 0.837 & \textbf{1.224} & \textbf{1.604} & \textbf{1.180} & \textbf{1.204} & 0.841 & 1.163 & 1.210 & \textbf{1.220} & 1.195 \\ \midrule
 
\multirow{2}{*}{ZH} & Mono & \textbf{0.833} & 1.101 & 1.142 & \textbf{1.012} & 0.995 & 0.816 & \textbf{1.093} & \textbf{1.156} & 0.990 & \textbf{1.004} \\
 & Multi & 0.819 & \textbf{1.137} & \textbf{1.271} & 0.990 & \textbf{1.001} & \textbf{0.825} & 1.088 & 1.000 & \textbf{1.041} & 0.999 \\ \midrule
 
\multirow{2}{*}{ES} & Mono & 0.849 & 1.024 & 1.046 & \textbf{1.148} & 1.150 & \textbf{0.857} & \textbf{1.235} & \textbf{1.176} & \textbf{1.141} & \textbf{1.182} \\
 & Multi & \textbf{0.852} & \textbf{1.146} & \textbf{1.214} & 1.130 & \textbf{1.152} & 0.849 & 1.082 & 1.055 & 1.129 & 1.148 \\ \midrule
 
 \multirow{2}{*}{FR} & Mono & 0.825 & 1.047 & 1.057 & \textbf{1.099} & 1.100 & 0.822 & \textbf{1.242} & \textbf{1.510} & \textbf{1.087} & 1.095 \\
 & Multi & \textbf{0.844} & \textbf{1.130} & \textbf{1.259} & 1.096 & \textbf{1.102} & \textbf{0.851} & 1.049 & 1.055 & 1.083 & \textbf{1.099} \\ \midrule

\multirow{2}{*}{HI} & Mono & \textbf{0.716} & 1.162 & \textbf{1.177} & 0.984 & 1.001 & 0.693 & \textbf{1.094} & \textbf{1.097} & 1.013 & \textbf{1.012} \\
 & Multi & 0.685 & \textbf{1.202} & 1.157 & \textbf{1.013} & 1.001 & \textbf{0.718} & 1.086 & 1.084 & \textbf{1.040} & 0.998 \\ \bottomrule
\end{tabular}%
%}
\caption{Predictive performance (Accuracy) and FA faithfulness (Suff, Comp, Soft-Suff, Soft-Comp) of mono- (BERT, RoBERTa) and multilingual models (mBERT, XLM-R). Full results per task and per FA are in Appendix \ref{app:full_results}.}
\label{tab:overview_results}

\end{table*}

\subsection{Predictive Performance} 

Table \ref{tab:overview_results} shows the predictive performance of models and faithfulness scores across FAs, averaged on three tasks for each language. Overall, we observe that the performance of mono- and multilingual models is consistently comparable to each other which demonstrates the importance of comparing FA faithfulness between these two types of models. 
For instance, the difference between Spanish BERT and mBERT is 0.003. The largest gap is found between Hindi BERT (0.716) and mBERT (0.685), exhibiting a difference of 0.031.

\subsection{Faithfulness} 

We note that FAs demonstrate inconsistent faithfulness discrepancies between mono- and multilingual models. In general, FAs obtain lower Suff and Comp with XLM-R than monolingual RoBERTa models, and higher with mBERT compared to monolingual BERT models (except for Comp in Hindi).
These discrepancies do not manifest though when we use Soft-Suff and Soft-Comp to measure faithfulness. In fact, the faithfulness of FAs between mono- and counterpart multilingual models are comparable to each other, i.e. the majority of differences are smaller than 0.01. For example, the greatest difference is only 0.051 (Soft-Comp between Chinese RoBERTa and XLM-R). 

Moreover, the faithfulness disparity in Suff and Comp of RoBERTa-based models is more noticeable as half of the cases have a faithfulness difference greater than 0.1. For example, the Comp in French is 1.51 for French RoBERTa but only 1.055 for XLM-R. We further investigate these disparities in faithfulness across metrics in Section \ref{sec:analysis}.

\begin{figure}[h]
\centering
\includegraphics[trim={0 .5cm 0 0}, width=0.99\columnwidth]{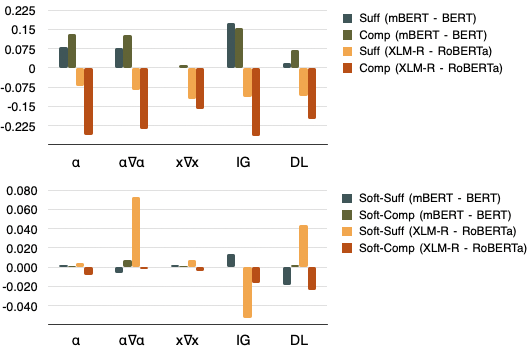}
\caption{Faithfulness disparity of FAs averaged across languages. Values above zero indicate that the FAs are more faithful in the multilingual model.}
\label{fig:agg_on_fa}
\end{figure}

\subsection{Comparing FAs}
Figure \ref{fig:agg_on_fa} delves deeper into the faithfulness disparity of FAs by looking at each one separately. Disparity is computed as the faithfulness score on the multilingual model minus the faithfulness score on the monolingual counterpart.\footnote{Tables \ref{tab:roberta}, \ref{tab:bert}, \ref{tab:soft_roberta}, and \ref{tab:soft_bert} in Appendix \ref{app:agg_results} show details per FA and language.}

First, Figure \ref{fig:agg_on_fa} (top) shows contrasting directions of faithfulness disparities between RoBERTa-based and BERT-based models. That is, FAs exhibit lower faithfulness in Suff and Comp for XLM-R than monolingual RoBERTa (above zero), whereas FAs exhibit higher Suff and Comp for mBERT than monolingual BERT. This holds true across FAs and languages as shown in Tables \ref{tab:bert} and \ref{tab:roberta} in Appendix \ref{app:agg_results}. 
Second, FAs do not show a consistent faithful disparity in terms of Soft-Suff and Soft-Comp (bottom) applied to mono- and multilingual models.

Moreover, we observe that IG has a larger faithfulness disparity than other FAs. For example, this is evident in Suff and Comp averaged over languages for both RoBERTa- and BERT-based models. IG is also the only FA showing significant differences in Suff and Comp disparity on both BERT- and RoBERTa-based models (see Table \ref{tab:bert} and Table \ref{tab:roberta} in Appendix).
This means, that when IG selects faithful rationales for a monolingual model, it might not be able to do so for the counterpart multilingual model. We also notice that the disparities of Attention-based FAs, i.e. $\alpha$ and $\alpha\nabla\alpha$, are consistently on par with each other.
However, they demonstrate larger disparities compared to  $x\nabla x$ and DL in most cases. Therefore, this indicates that mono- and multilingual models tend to employ attention differently for making predictions.

On the other hand, the results of Soft-Suff and Soft-Comp do not echo any of the above observations we made for Suff and Comp. Indeed, the Soft-Suff and Soft-Comp disparities of FAs are not significantly different between mono- and multilingual models (see p-values in Appendix \ref{app:agg_results}). That is, when evaluating the overall importance distribution given by a FA, there is no faithfulness discrepancy between using a multilingual or a monolingual model. One possible reason is that, the pre-defined rationale lengths introduce bias to the evaluation of Suff and Comp. To investigate this, we examine the faithfulness disparity on specific rationale lengths. We find that the faithfulness disparity varies across rationale lengths. Table \ref{tab:faithfulness_at_02} shows the Suff disparity at different rationale lengths, 10\% and 50\%, within BERT-based models. For example, FAs are more faithful with XLM-R than Spanish RoBERTa at a 10\% rationale length, with a significant average difference of 0.236 (p-value of 0.03) across tasks. When looking at the rationale length of 50\%, FAs are comparably faithful for XLM-R and Spanish RoBERTa (avg of 0.027, p-value of 0.245). This further inspires us to consider the impact of the tokenizers on faithfulness. Soft-Suff and Soft-Comp evaluate the importance scores over the whole input and this can be less sensitive to the tokenization. We examine this in Section \ref{sec:analysis}.

\section{Analysis}
\label{sec:analysis}

We further investigate if the tokenization and model size contribute to the contrasting directions of FA faithfulness disparity between BERT- and RoBERTa-based models.

\UseRawInputEncoding
\newcolumntype{R}{>{\testnext}r<{}}
%%%%%%% bert soft
\begin{table}[!h]
\centering
\resizebox{0.99\columnwidth}{!}{%
\begin{tabular}{@{}c|RRRRRRccccc|cc|@{}}
\hline
\multicolumn{8}{c}{\textbf{Sufficiency at 10\%}} \\     
\hline
\toprule
         & $\alpha$ & $\alpha\nabla\alpha$ & $x\nabla x$ & IG     & DL      & Avg Diff & P value \\ \midrule
English & 0.225 & 0.232 & -0.021 & 0.168 & 0.035 & 0.128 & 0.202 \\
Chinese & 0.037 & 0.007 & 0.079 & 0.285 & 0.054 & 0.093 & 0.228 \\
Spanish & 0.41 & 0.407 & 0.084 & 0.346 & -0.067 & 0.236 & 0.03 \\
French & 0.256 & 0.236 & -0.011 & 0.21 & -0.047 & 0.129 & 0.042 \\
Hindi & -0.198 & -0.185 & -0.06 & 0.244 & -0.075 & -0.055 & 0.575 \\ \midrule
Avg Diff & 0.146 & 0.139 & 0.014 & 0.251 & -0.02 & 0.106 &  \\
P value & \text{0.224} & \text{0.255} & \text{0.747} & \text{0.004} & \text{0.666} &  & \text{0.008} \\
\hline
\multicolumn{8}{c}{\textbf{Sufficiency at 50\%}} \\     
\hline
\toprule
         & $\alpha$ & $\alpha\nabla\alpha$ & $x\nabla x$ & IG     & DL      & Avg Diff & P value \\ \midrule
English & -0.146 & -0.15 & -0.022 & 0.05 & 0.058 & -0.042 & 0.3 \\
Chinese & 0.061 & 0.062 & 0.045 & 0.15 & -0.042 & 0.055 & 0.271 \\
Spanish & 0.033 & 0.035 & -0.026 & 0.075 & 0.019 & 0.027 & 0.245 \\
French & 0.096 & 0.09 & -0.021 & -0.046 & 0.059 & 0.036 & 0.313 \\
Hindi & 0.026 & 0.031 & 0.049 & 0.251 & 0.02 & 0.075 & 0.094 \\ \midrule
Avg Diff & 0.014 & 0.014 & 0.005 & 0.096 & 0.023 & 0.03 &  \\
P value & \text{0.752} & \text{0.765} & \text{0.831} & \text{0.039} & \text{0.51} &  & \text{0.082}\\
\bottomrule
\end{tabular}%
}
\caption{Sufficiency difference between mBERT and counterpart monolingual BERT on rationale ratio of 10\% and 50\%. \textcolor{purple}{Plum} indicates that monolingual models are more faithful than multilingual models.
}
\label{tab:faithfulness_at_02}
\end{table}

\subsection{Impact of model size}\label{sec:impact_of_model_size}

We posit that the difference between RoBERTa and BERT-based models in disparity directions of FAs faithfulness is associated with the differences in model sizes of mono- and multilingual models.
Specifically, mBERT has at least 1.5 times more parameters than monolingual BERT models, while XLM-R has at least 2.2 times more parameters than monolingual RoBERTa models. The difference in model size may account for the opposite directions of faithfulness disparities between RoBERTa- and BERT-based models.
If this holds true, we anticipate that when the model size gap increases, XLM-R will still provide less faithful rationales than monolingual RoBERTa while their disparity degree will increase. 

To further investigate the impact of the model size, we repeat all experiments using XLM-R large and compare its faithfulness with monolingual RoBERTa. Full results are in Table \ref{tab:roberta_large} and Table \ref{tab:full_result_robert_large} in Appendix. In this case, we examine a RoBERTa-based model pair with a larger difference in model size than XLM-R base vs. monolingual RoBERTa. XLM-R base and XLM-R large use the same pre-training corpus, pre-training objective, and similar model architectures, but differ in model parameter numbers\footnote{Both are transformer-based, XLM-R base: L = 12, H = 768, A = 12; XLM-R large: L = 24, H = 1024, A = 16)} \cite{conneau-etal-2020-unsupervised}. XLM-R large (550M parameters) is at least 4.7 times larger than the monolingual RoBERTa models.

The results first show that the faithfulness disparity direction remains the same as the one between XLM-R base and monolingual RoBERTa. This implies that FAs are more faithful with monolingual RoBERTa. Second, the overall sufficiency disparity increases from -0.100 to -0.186. It also increases for each individual FA and language, with IG being the only exception by remaining almost the same (-0.120 and -0.121). For example, the average disparity in English increases from -0.143 to -0.300 and the average disparity for attention increases from -0.070 to -0.195. 
The overall comprehensiveness disparity of XLM-R large is on par with XLM-R base (-0.226 v.s. -0.197).  

Overall, the results confirm our assumption that the difference in model size is related to the faithfulness disparity. The larger the multilingual model, the less faithful its rationales are compared to its monolingual counterpart. One intuitive interpretation behind this is that when the model gets larger, it becomes intrinsically complex and therefore, it is harder to faithfully explain its predictions with FA. To summarize, \textit{the more parameters the multilingual model has, the less faithful its rationales are compared to its monolingual counterparts.} 

We acknowledge that our findings might not generalize to BERT because mBERT models of different sizes are not available to experiment with. To overcome this, we repeat all experiments with BERT-large and compare its faithfulness with BERT-base, to investigate the impact of model size from a different perspective. The results show that FAs obtain lower Suff and Comp with the larger BERT model across FAs and tasks. This observation is in agreement with our assumption above that model sizes might impact faithfulness disparity. To keep the focus of the paper on the faithfulness disparity between mono- and multilingual models, we present the results and analysis in Table \ref{tab:added_bert_base_large_comparision} in the Appendix.

\subsection{Impact of tokenization}\label{sec:impact_of_tokenization}

\begin{table*}[]
\centering
\resizebox{0.88\textwidth}{!}{%
\begin{tabular}{lcccccccc}
\hline
\multicolumn{9}{c}{\textbf{RoBERTa}} \\ \hline
\multicolumn{1}{l|}{} & \multicolumn{1}{l}{Multi Fertility} & \multicolumn{1}{l}{Mono Fertility} & \multicolumn{1}{l|}{Fertility Diff} & \multicolumn{1}{l}{Multi Splitting} & \multicolumn{1}{l}{Mono Splitting} & \multicolumn{1}{l|}{Split ratio Diff} & \multicolumn{1}{l}{Suff Diff} & \multicolumn{1}{l}{Comp Diff} \\ \hline
\multicolumn{1}{l|}{English} & 1.179 & 1.115 & \multicolumn{1}{r|}{0.064} & 0.111 & 0.059 & \multicolumn{1}{r|}{0.052} & 0.078 & 0.079 \\
\multicolumn{1}{l|}{Spanish} & 1.369 & 1.283 & \multicolumn{1}{r|}{0.086} & 0.152 & 0.090 & \multicolumn{1}{r|}{0.062} & 0.123 & 0.168 \\
\multicolumn{1}{l|}{French} & 1.461 & 1.456 & \multicolumn{1}{r|}{0.005} & 0.139 & 0.134 & \multicolumn{1}{r|}{0.005} & 0.083 & 0.202 \\ \hline
\multicolumn{1}{l|}{Avg} & 1.336& 1.285& \multicolumn{1}{r|}{0.052} & 0.134& 0.094& \multicolumn{1}{r|}{0.040} & 0.095& 0.150\\ \hline

\multicolumn{9}{c}{\textbf{BERT}} \\ \hline
\multicolumn{1}{l|}{} & \multicolumn{1}{l}{Multi Fertility} & \multicolumn{1}{l}{Mono Fertility} & \multicolumn{1}{l|}{Fertility Diff} & \multicolumn{1}{l}{Multi Splitting} & \multicolumn{1}{l}{Mono Splitting} & \multicolumn{1}{l|}{Splitting Diff} & \multicolumn{1}{l}{Suff Diff} & \multicolumn{1}{l}{Comp Diff} \\ \hline
\multicolumn{1}{l|}{English} & 1.319 & 1.125 & \multicolumn{1}{r|}{0.195} & \underline{0.261} & \underline{0.076} & \multicolumn{1}{r|}{0.185} & -0.300 & -0.250 \\
\multicolumn{1}{l|}{Spanish} & 1.409 & 1.290 & \multicolumn{1}{r|}{0.119} & 0.299 & 0.195 & \multicolumn{1}{r|}{0.104} & -0.240 & -0.099 \\
\multicolumn{1}{l|}{French} & 1.531 & 1.345 & \multicolumn{1}{r|}{0.186} & 0.325 & 0.211 & \multicolumn{1}{r|}{0.114} & -0.236 & -0.434 \\ \hline
\multicolumn{1}{l|}{Avg} & 1.420& 1.253& \multicolumn{1}{r|}{0.167} & 0.312& 0.203& \multicolumn{1}{r|}{0.134} & -0.259& -0.261\\ \hline
\end{tabular}%
}
\caption{Fertility, splitting ratio, sufficiency, and comprehensiveness difference between multilingual (``Multi Fertility'', ``Multi Splitting'') and monolingual models (``Mono Fertility'', ``Mono Splitting''). For ``Suff Diff'' and ``Comp Diff'', positive values indicate that the FA is more faithful to the multilingual model. Full results of fertility and splitting ratio for each dataset can be found in Table \ref{tab:fertility_full_results} in Appendix \ref{app:fertility_full_results}.}
\label{tab:tokenizer_diff}
\end{table*}

Previous research has shown the impact of tokenization on multilingual models \cite{ruan-etal-2021-sattiy,zhang-etal-2022-robust}. Intuitively, multilingual tokenizers are less specialized than their counterpart monolingual tokenizers for the specific language. For example, the multilingual BERT tokenizer has a vocabulary size of 105K covering 104 languages, while the five monolingual BERT tokenizers cover a vocabulary of 167k tokens (see Table \ref{table:models}). 
BERT-based models use WordPiece as their tokenizers \citep{wu2016google}. Monolingual RoBERTa models use BytePair-Encoding (BPE) \citep{sennrich-etal-2016-neural}, and the multilingual XLM-R uses SentencePiece \citep{kudo-richardson-2018-sentencepiece}. 

Therefore, we investigate the impact of tokenization on faithfulness disparity. 
The effectiveness of a tokenizer in text splitting intuitively reflects how many unique tokens it knows in a particular language. Following \citet{rust-etal-2021-good}, we examine two metrics for assessing tokenization, fertility and splitting ratio. Fertility indicates how many subwords a tokenizer splits a word into, while the splitting ratio shows how often a tokenizer splits words. Intuitively, low scores are preferable for both metrics indicating that the tokenizer is well-suited to the language \cite{rust-etal-2021-good}.

Table \ref{tab:tokenizer_diff} shows the fertility and splitting ratio difference between monolingual and multilingual models (i.e. multilingual score minus its counterpart monolingual).\footnote{Hindi and Chinese are excluded from this analysis. For Hindi, we do not observe a substantial difference between mono- and multilingual models in Suff or Comp. Chinese is a logographic language without white spaces. For reference, we present the fertility and splitting ratios for Hindi in Table~\ref{tab:hindi_split_ratio} in the Appendix.} Faithfulness disparity values are taken from Tables \ref{tab:bert} and \ref{tab:roberta}.

First, for both RoBERTa and BERT-based models, the positive values of fertility and splitting ratio difference indicate that multilingual models tend to be more aggressive in splitting words than monolingual ones. For example, as shown in Table \ref{tab:fertility_full_results} in Appendix \ref{app:fertility_full_results}, 26.1\% English words (underlined in table) are split by the SentencePiece tokenizer of XLM-R but only 7.6\% (underlined in table) by BPE which is used in monolingual RoBERTa models.

Second, RoBERTa-based models have larger gaps in both fertility and splitting ratio than BERT-based across all three languages. The fertility and the splitting ratio differences are greater than 0.1 for RoBERTa-based, but less than 0.1 for BERT-based models. This is because SentencePiece (multilingual XLM-R's tokenizer) is generally more aggressive in splitting words. Taking English as an example, the fertility gaps among monolingual RoBERTa (BPE), monolingual BERT (WordPiece) tokenizers, and multilingual BERT (WordPiece) are relatively smaller, 1.125, 1.115, and 1.179 respectively, while the fertility of XLM-R (SentencePiece) is 1.319. However, this is counterintuitive given the much larger vocabulary size of XLM-R, over two times bigger than multilingual BERT (see Figure \ref{table:models}). One potential explanation is that XLM-R saves capacity for representing the vocabulary for other low-resource languages. 
On the other hand, the greater aggressiveness in tokenization of XLM-R potentially explains the different disparity direction in Suff and Comp to BERT models. That is, only when the fertility difference is greater than 0.1, FAs are more faithful with multilingual models than with monolingual counterparts.\footnote{We further demonstrate this pattern in Figure \ref{fig:SentencePiece} in Appendix \ref{app:SentencePiece}.}

\begin{table}[!h]
\centering
\resizebox{0.48\columnwidth}{!}{%
\begin{tabular}{@{}r|cc@{}}
\toprule
$\rho$ & \multicolumn{1}{l}{Suff Diff} & \multicolumn{1}{l}{Comp Diff} \\ \midrule
Splitting Diff & -0.86 & -0.79 \\
Fertility Diff & -0.86 & -0.91 \\ \bottomrule
\end{tabular}%
}
\caption{Pearson correlation coefficient between fertility, splitting ratio, and faithfulness disparity.}
\label{tab:cc}
\end{table}

Last, as shown in Table~\ref{tab:cc}, the differences in Suff and Comp demonstrate a highly negative relationship to the fertility difference. That is, the larger the fertility difference between mono- and multilingual models, the smaller the faithfulness disparity. Particularly, the fertility and the comprehensiveness difference show a very high negative correlation (-0.91).

To sum up, \textit{multilingual tokenizers split words into subwords more aggressively than monolingual tokenizers, which potentially contributes to the faithfulness disparity between models. The aggressive tokenization of multilingual models might result in lower faithfulness.}

\subsection{Disentangling the impact of the model}\label{sec:disentangle_impact_of_models}

\begin{table*}[!t]
\centering
\resizebox{0.79\textwidth}{!}{%
\begin{tabular}{l|ccccc|ccccc}
\hline
\multicolumn{1}{c|}{} & \multicolumn{5}{c|}{\textbf{Sufficiency}} & \multicolumn{5}{c}{\textbf{Soft Sufficiency}} \\ \hline
\multicolumn{1}{c|}{} & $\alpha$ & $\alpha\nabla\alpha$ & $x\nabla x$ & IG & DL & $\alpha$ & $\alpha\nabla\alpha$ & $x\nabla x$ & IG & DL \\ \hline
RoBERTa (FR) & 1.287 & 1.289 & 1.198 & 1.232 & 1.201 & 1.091 & 1.033 & 1.106 & 1.087 & 1.120 \\
RoBERTa (EN $\rightarrow$ FR) & 1.241 & 1.243 & 1.193 & 1.225 & 1.197 & 1.090 & 1.029 & 1.098 & 1.083 & 1.122 \\
XLM-R (Multi $\rightarrow$ FR) & 1.230 & 1.200 & 1.242 & 1.246 & 1.032 & 1.088 & 1.032 & 1.102 & 1.086 & 1.120 \\
XLM-R & 1.081 & 1.071 & 1.065 & 1.015 & 1.013 & 1.046 & 1.144 & 1.012 & 1.100 & 1.114 \\ \hline
\multicolumn{1}{c|}{} & \multicolumn{5}{c|}{\textbf{Comprehensiveness}} & \multicolumn{5}{c}{\textbf{Soft Comprehensiveness}} \\ \hline
\multicolumn{1}{c|}{} & $\alpha$ & $\alpha\nabla\alpha$ & $x\nabla x$ & IG & DL & $\alpha$ & $\alpha\nabla\alpha$ & $x\nabla x$ & IG & DL \\ \hline
RoBERTa (FR) & 1.573 & 1.567 & 1.267 & 1.667 & 1.476 & 1.097 & 1.093 & 1.090 & 1.102 & 1.092 \\
RoBERTa (EN $\rightarrow$ FR) & 1.305 & 1.317 & 1.321 & 1.474 & 1.309 & 1.099 & 1.09 & 1.087 & 1.099 & 1.091 \\
XLM-R (Multi $\rightarrow$ FR) & 1.394 & 1.401 & 1.266 & 1.435 & 1.353 & 1.097 & 1.090 & 1.089 & 1.100 & 1.091 \\
XLM-R & 1.087 & 1.085 & 1.035 & 1.069 & 1.001 & 1.100 & 1.097 & 1.099 & 1.101 & 1.097 \\ \hline
\end{tabular}%
}
\caption{Faithfulness of French RoBERTa, XLM-R and two adapted models to French averaged over French tasks. }
\label{tab:french_monoModel_multitokenizer_mix}
\end{table*}

To further investigate how tokenization and model selection affects faithfulness, we experiment with (1) adapting a monolingual model to a different language (i.e. EN to FR); and (2) adapting a multilingual model to FR. This allows us to disentangle the impact of the model itself while observing the faithfulness changes of FAs across tokenization strategies. We experiment with RoBERTa-based models (RoBERTa and XLM-R) in French because this was the case where we observed the greatest faithfulness discrepancy between a mono- and multilingual model (see Table \ref{tab:roberta}).

We use WECHSEL \citep{minixhofer-etal-2022-wechsel} to adapt an English RoBERTa to French. We replace the tokenizer of the English RoBERTa with the tokenizer of the French RoBERTa. French token embeddings are initialized such that they are semantically similar to the English tokens by using multilingual static word embeddings covering English and French. We refer to this model as \textbf{RoBERTa (EN $\rightarrow$ FR)}. 
For monolingual specialization (i.e. French) of XLM-R, we use FOCUS \citep{dobler-demelo-2023-focus} to replace the tokenizer of XLM-R with the tokenizer of French RoBERTa. FOCUS first finds the shared tokens between the French RoBERTa and XLM-R vocabularies which XLM-R can use directly. New tokens (French tokens not in XLM-R) are represented as combinations of overlapping tokens in the French RoBERTa and XLM-R vocabularies. We refer to this model as \textbf{XLM-R (Multi $\rightarrow$ FR)}.\footnote{See Appendix \ref{app:french_focus_implementation} for the implementation details of these two models.} 

The first three models in Table \ref{tab:french_monoModel_multitokenizer_mix}, namely French RoBERTa (FR), RoBERTa (EN $\rightarrow$ FR) and XLM-R (Multi $\rightarrow$ FR), use the same tokenizer, i.e. the same tokenization aggressiveness, while the embedding initialization is either identical or semantically similar \citep{minixhofer-etal-2022-wechsel,dobler-demelo-2023-focus}. Notably, all FAs obtain more similar Suff and Comp between RoBERTa (FR) and the two hybrid RoBERTa, rather than between XLM-R and the two hybrid RoBERTa. For example, each FA on XLM-R (Multi $\rightarrow$ FR) almost mirrors the sufficiency of RoBERTa (FR) with the greatest difference of 0.232 and the smallest of 0, even though XLM-R (Multi $\rightarrow$ FR) shares the same model parameters (non-embedding weights) with XLM-R (their greatest and smallest differences are 0.366 and 0.001, respectively). 
Therefore, we summarize that \textit{FAs tend to be of similar faithfulness on models with the same tokenizer or the similar tokenizers regarding the splitting aggressiveness level.} Further, this leads to promising future research questions. First, do FAs really reflect the inner reasoning process of models? Second, when token units and their embeddings are identical or similar, different models tend to converge to a point after fine-tuning where they process these inputs in a similar way?

\subsection{Qualitative analysis}\label{sec:quali}

For a qualitative evaluation, we examine the rationales extracted by the same FAs for both types of models. 
We observe that rationales of multilingual models more often contain pronouns, prepositions, postpositions, conjunction, and article words, while monolingual models' prefer nouns and adjectives. We suspect the different preferences in parts of speech are due to monolingual models being more specialized for the language so that its rationales contain more specific nouns and adjectives rather than general functional words such as pronouns, prepositions, postpositions, and conjunctions. 

We also observe examples where multilingual tokenizers split text more aggressively than monolingual tokenizers, e.g. the word ``defectos'' in Spanish (``defects'' in English) is not split into subwords by Spanish BERT, but split into `def', `\#\#ecto', `\#\#s' by mBERT; ``desagradable'' in Spanish (``unpleasant'' in English) is not split by Spanish BERT but split into `desa', `\#\#grada', `\#\#ble' by mBERT, echoing the observations in Section \ref{sec:impact_of_tokenization}.

\section{Conclusion}

To the best of our knowledge, our study is the first to investigate the faithfulness disparity between monolingual and multilingual models. We have conducted a comprehensive empirical study and found that faithfulness gaps exist across languages, models, and FAs. 
Our study further reveals that the larger the multilingual model, the less faithful its rationales are compared to its monolingual counterpart models.
Finally, we found that the disparity is highly correlated to the gap between mono- and multilingual tokenizers on how aggressively they split words. Further experiments support the assumption on the impact of tokenization: the discrepancies in faithfulness are primarily driven by differences in tokenization rather than underlying differences in how mono- and multilingual models semantically process the input.

\section*{Limitations}
A significant challenge we encountered during our research was the absence of monolingual models in various languages. First, monolingual models are only available in a few languages, such as monolingual BERT and RoBERTa models used in this paper. Second, more recent decoder-based models, such as Llama~\citep{touvron2023llama}, Mistral~\citep{jiang2023mistral}, and Gemma~\citep{team2024gemma}, are multilingual by default. Furthermore, it would be intriguing to explore the faithfulness disparity and behavior of feature attributions for low-resource languages, particularly given their limited presence in the pre-training corpora.

Finally, it is important to acknowledge that multilingual studies focusing on Indo-European and Sino-Tibetan languages may not necessarily apply to languages outside these language families. We hope future work can contribute resources to facilitate the development of a more diverse range of monolingual language models.

\section*{Acknowledgements}

ZZ and NA are supported by EPSRC grant EP/V055712/1, part of the European Commission CHIST-ERA programme, call 2019 XAI: Explainable Machine Learning-based Artificial Intelligence. We thank George Chrysostomou for his invaluable feedback.

\bibliography{anthology,custom}
\bibliographystyle{acl_natbib}

\clearpage

%%%%%%%%%%% APP
\appendix

\section{Comparison of predictive performance}

\begin{table}[!h]
\resizebox{\columnwidth}{!}{%
\begin{tabular}{@{}lllllll@{}}
\toprule
{  } & {  } & {  \textbf{NER}} & {  SA} & {  \textbf{QA}} & {  \textbf{UDP}} & {  \textbf{POS}} \\
\multirow{-2}{*}{{  \textbf{Lg}}} & \multirow{-2}{*}{{  \textbf{Model}}} & Test F1 & Test Acc & Dev EM / F1 & Test UAS/LAS & Test Acc \\ \midrule
Arabic & {  Monolingual} & {  91.1} & {  95.9} & {  68.3/82.4} & {  90.1/85.6} & {  96.8} \\
{  AR} & {  mBERT} & {  90} & {  95.4} & {  66.1/80.6} & {  88.8/83.8} & {  96.8} \\ \midrule
 & {  Monolingual} & {  91.5} & {  91.6} & {  80.5/88.0} & {  92.1/89.7} & {  97} \\
\multirow{-2}{*}{English} & {  mBERT} & {  91.2} & {  89.8} & {  80.9/88.4} & {  91.6/89.1} & {  96.9} \\ \midrule
 & {  Monolingual} & {  92} & - & {  69.9/81.6} & {  95.9/94.4} & {  98.4} \\
\multirow{-2}{*}{Finnish} & {  mBERT} & {  88.2} & - & {  66.6/77.6} & {  91.9/88.7} & {  96.2} \\ \midrule
 & {  Monolingual} & {  91} & {  96} & {  66.8/78.1} & {  85.3/78.1} & {  92.1} \\
\multirow{-2}{*}{Indonesian} & {  mBERT} & {  93.5} & {  91.4} & {  71.2/82.1} & {  85.9/79.3} & {  93.5} \\ \midrule
 & {  Monolingual} & {  72.4} & {  88} & - & {  94.7/93.0} & {  98.1} \\
\multirow{-2}{*}{Japanese} & {  mBERT} & {  73.4} & {  87.8} & - & {  94.0/92.3} & {  97.8} \\ \midrule
 & {  Monolingual} & {  88.8} & {  89.7} & {  74.2/91.1} & {  90.3/87.2} & {  97} \\
\multirow{-2}{*}{Korean} & {  mBERT} & {  86.6} & {  86.7} & {  69.7/89.5} & {  89.2/85.7} & {  96} \\ \midrule
 & {  Monolingual} & {  91} & {  95.2} & {  64.3/83.7} & {  93.1/89.9} & {  98.4} \\
\multirow{-2}{*}{Russian} & {  mBERT} & {  90} & {  95} & {  63.3/82.6} & {  91.9/88.5} & {  98.2} \\ \midrule
 & {  Monolingual} & {  92.8} & {  88.8} & {  60.6/78.1} & {  79.8/73.2} & {  96.9} \\
\multirow{-2}{*}{Turkish} & {  mBERT} & {  93.8} & {  86.4} & {  57.9/76.4} & {  74.5/67.4} & {  95.7} \\ \midrule
{  } & {  Monolingual} & {  76.5} & {  95.3} & {  82.3/89.3} & {  88.6/85.6} & {  97.2} \\
\multirow{-2}{*}{{  Chinese}} & {  mBERT} & {  76.1} & {  93.8} & {  82.0/89.3} & {  88.1/85.0} & {  96.7} \\ \midrule
{  } & {  Monolingual} & {  87.4} & {  92.4} & {  70.8/84.0} & {  90.0/86.3} & {  96.9} \\
\multirow{-2}{*}{{  AVG}} & {  mBERT} & {  87} & {  91} & {  69.7/83.3} & {  88.4/84.4} & {  96.4} \\ \bottomrule
\end{tabular}%
}
\caption{Comparison of predictive performance between mBERT and monolingual BERT across languages and tasks. Results are drawn from \citet{rust-etal-2021-good}}
\label{tab:how_good_is}
\end{table}

As shown in Table \ref{tab:how_good_is}, the predictive performance of mBERT is comparable to monolingual BERT in most cases. Particularly, the difference between monolingual and multilingual models is not greater than 1.2 and 1.5 across each task in Russian and Chinese respectively.

\section{Model Implementation Details}
\label{app:model_hyperparameters}

\begin{table}[hbt!]
\resizebox{\columnwidth}{!}{%
\begin{tabular}{@{}llll@{}}
\toprule
\textbf{Language} & \textbf{Models} & \textbf{Huggingface ID} &  \\ \midrule
\multirow{2}{*}{\textbf{Multilingual}} & mBERT & bert-base-multilingual-uncased & \citet{devlin-etal-2019-bert} \\
 & XLM-R & xlm-roberta-base & \citet{conneau-etal-2020-unsupervised} \\ 
  & XLM-R large & xlm-roberta-large & \citet{conneau-etal-2020-unsupervised} \\ \midrule
\multirow{2}{*}{\textbf{English}} & BERT & bert-base-uncased & \citet{devlin-etal-2019-bert} \\
 & RoBERTa & roberta-base & \citet{liu2019roberta} \\ \midrule
\multirow{2}{*}{\textbf{Chinese}} & BERT & bert-base-chinese & \citet{devlin-etal-2019-bert} \\
 & RoBERTa & hfl/chinese-roberta-wwm-ext & \citet{cui2021pre} \\ \midrule
\multirow{2}{*}{\textbf{Spanish}} & BERT & dccuchile/bert-base-spanish-wwm-uncased & \citet{CaneteCFP2020} \\
 & RoBERTa & PlanTL-GOB-ES/roberta-base-bne & \citet{gutierrez2021maria} \\ \midrule
\multirow{2}{*}{\textbf{French}} & BERT & dbmdz/bert-base-french-europeana-cased & \citet{stefan_schweter_2020_4275044} \\
 & RoBERTa & ClassCat/roberta-base-french & n/a \\ \midrule
\multirow{2}{*}{\textbf{Hindi}} & BERT & l3cube-pune/hindi-bert-scratch & \citet{joshi2022l3cubehind} \\
 & RoBERTa & flax-community/roberta-hindi & n/a \\ \bottomrule
\end{tabular}%
}
\caption{Model references}
\label{tab:model_references}
\end{table}

We use pre-trained models from the Huggingface library \citep{wolf-etal-2020-transformers}. We use the AdamW optimizer \citep{loshchilov2017decoupled} with a learning rate of $1e^{-5}$ for fine-tuning ( $1e^{-4}$ for the linear output layer). We fine-tune all models for five epochs using a linear scheduler, with 10\% of the data in the first epoch as warming up. We also use a grad-norm of 1.0. The model with the lowest loss on the development set is selected. All models are trained across three random seeds, and we report the average prediction performance. The best model among the three runs is used to extract rationales. Experiments are run on a single NVIDIA A100 GPU.

\section{Faithfulness disparity on FAs and languages}\label{app:agg_results}

\newcolumntype{R}{>{\testnext}r<{}}
\begin{table}[!h]
\centering
\resizebox{0.95\columnwidth}{!}{%
\begin{tabular}{@{}c|RRRRRRccccc|cc|@{}}
\hline
\multicolumn{8}{c}{\textbf{Sufficiency}} \\     
\hline
\toprule
         & $\alpha$ & $\alpha\nabla\alpha$ & $x\nabla x$ & IG     & DL      & Avg Diff & P value \\ \midrule
English & 0.086 & 0.093 & -0.024 & 0.187 & 0.048 & 0.078 & 0.292 \\
Chinese & -0.018 & -0.037 & 0.043 & 0.176 & 0.016 & 0.036 & 0.454 \\
Spanish & 0.200 & 0.202 & 0.006 & 0.190 & 0.015 & 0.123 & 0.049 \\
French & 0.184 & 0.173 & -0.028 & 0.063 & 0.025 & 0.083 & 0.066 \\
Hindi & -0.041 & -0.035 & 0.010 & 0.266 & -0.003 & 0.039 & 0.510 \\
\midrule
Avg Diff & 0.082 & 0.079 & 0.001 & 0.176 & 0.020 & 0.072 & - \\
P value  & \text{0.264}  & \text{0.298}  & \text{0.966}  & \text{0.003} & \text{0.527}  & \text{-}   & \text{0.005}  \\ \hline
\hline
\multicolumn{8}{c}{\textbf{Comprehensiveness}} \\     
\hline
\toprule
         & $\alpha$ & $\alpha\nabla\alpha$ & $x\nabla x$ & IG     & DL      & Avg Diff & P value \\ \midrule
English & 0.122 & 0.106 & 0.075 & 0.078 & 0.015 & 0.079 & 0.323 \\
Chinese & 0.211 & 0.213 & 0.028 & 0.176 & 0.016 & 0.129 & 0.053 \\
Spanish & 0.268 & 0.268 & 0.040 & 0.160 & 0.105 & 0.168 & 0.048 \\
French & 0.294 & 0.299 & 0.046 & 0.217 & 0.156 & 0.202 & 0.049 \\
Hindi & -0.232 & -0.234 & -0.128 & 0.138 & 0.057 & -0.080 & 0.307 \\ \midrule
Avg Diff & 0.133 & 0.130 & 0.012 & 0.154 & 0.070 & 0.100 & -   \\
P value  & \text{0.258}  & \text{0.263}  & \text{0.758}  & \text{0.040} & \text{0.081} & \text{-}   & \text{0.007}  \\ \bottomrule

\end{tabular}%
}
\caption{Suff and Comp difference between multilingual BERT (mBERT) and monolingual BERT. ({\small \textcolor{purple}{plum} indicates monolingual models are more faithful than multilingual models}.}
\label{tab:bert}
\end{table}

\newcolumntype{R}{>{\testnext}r<{}}

\begin{table}[!ht]
\centering
\resizebox{0.95\columnwidth}{!}{%
\begin{tabular}{@{}c|RRRRRRccccc|cc|@{}}
\hline
\multicolumn{8}{c}{\textbf{Sufficiency}} \\     
\hline
\toprule
         & $\alpha$ & $\alpha\nabla\alpha$ & $x\nabla x$ & IG     & DL      & Avg Diff & P value \\ \midrule
English & -0.082 & -0.086 & -0.097 & -0.131 & -0.319 & -0.143 & 0.258 \\
Chinese & 0.065 & 0.056 & -0.085 & -0.040 & -0.018 & -0.005 & 0.946 \\
Spanish & -0.070 & -0.138 & -0.336 & -0.107 & -0.111 & -0.153 & 0.053 \\
French & -0.206 & -0.218 & -0.133 & -0.217 & -0.188 & -0.193 & 0.007 \\
Hindi & -0.054 & -0.047 & 0.045 & -0.068 & 0.081 & -0.009 & 0.888 \\ \midrule
Avg Diff & -0.070 & -0.086 & -0.121 & -0.113 & -0.111 & -0.100 & -   \\
P value  & \text{0.535}  & \text{0.462}  & \text{0.041}  & \text{0.033}  & \text{0.076}  & \text{-}  & \text{0.006}   \\ \hline
\hline
\multicolumn{8}{c}{\textbf{Comprehensiveness}} \\     
\hline
\toprule
         & $\alpha$ & $\alpha\nabla\alpha$ & $x\nabla x$ & IG     & DL      & Avg Diff & P value \\ \midrule
English & -0.465 & -0.436 & -0.327 & -0.333 & -0.330 & -0.378 & 0.000 \\
Chinese & -0.230 & -0.224 & -0.111 & -0.156 & -0.062 & -0.157 & 0.010 \\
Spanish & -0.197 & -0.116 & -0.105 & 0.032 & -0.218 & -0.121 & 0.076 \\
French & -0.486 & -0.482 & -0.232 & -0.598 & -0.475 & -0.455 & 0.004 \\
Hindi & 0.071 & 0.062 & -0.036 & -0.268 & 0.082 & -0.018 & 0.831 \\ \midrule
Avg Diff & -0.261 & -0.239 & -0.162 & -0.265 & -0.201 & -0.226 & -   \\
P value  & \text{0.027}  & \text{0.034}  & \text{0.004}  & \text{0.015}  & \text{0.070}  & \text{-}  & \text{0.000}  \\ \bottomrule
\end{tabular}%
}
\caption{Suff and Comp difference between XLM-R (multilingual RoBERTa) and monolingual RoBERTa ({\small \textcolor{purple}{plum} indicates monolingual models are more faithful than multilingual models}.)
}
\label{tab:roberta}
\end{table}

\begin{table}[!ht]
\centering
\resizebox{0.95\columnwidth}{!}{%
\begin{tabular}{@{}c|RRRRRRccccc|cc|@{}}
\hline
\multicolumn{8}{c}{\textbf{Soft-Sufficiency}} \\     
\hline
\toprule
         & $\alpha$ & $\alpha\nabla\alpha$ & $x\nabla x$ & IG     & DL      & Avg Diff & P value \\ \midrule
English & -0.018 & 0.048 & 0.015 & 0.033 & -0.038 & 0.008 & 0.821 \\
Chinese & 0.024 & 0.021 & -0.073 & -0.036 & -0.046 & -0.022 & 0.45 \\
Spanish & -0.105 & 0.027 & 0.045 & -0.028 & -0.032 & -0.019 & 0.539 \\
French & 0.06 & -0.039 & -0.011 & 0.043 & -0.069 & -0.003 & 0.915 \\
Hindi & 0.05 & -0.085 & 0.035 & 0.058 & 0.088 & 0.029 & 0.284 \\
\midrule
Avg Diff & 0.002 & -0.006 & 0.002 & 0.014 & -0.019 & -0.001 & - \\
P value & \text{0.931} & \text{0.869} & \text{0.942} & \text{0.635} & \text{0.499} & - & \text{0.92} \\
\hline
\hline
\multicolumn{8}{c}{\textbf{Soft-Comprehensiveness}} \\     
\hline
\toprule
         & $\alpha$ & $\alpha\nabla\alpha$ & $x\nabla x$ & IG     & DL      & Avg Diff & P value \\ \midrule
English & 0.003 & 0.003 & \text{0} & \text{0} & 0.004 & 0.002 & 0.273 \\
Chinese & -0.001 & 0.034 & 0.001 & -0.006 & 0.001 & 0.006 & 0.408 \\
Spanish & 0.003 & 0.002 & 0.001 & 0.002 & 0.004 & 0.002 & 0.036 \\
French & 0 & 0.004 & 0.001 & 0.004 & 0.003 & 0.002 & 0.041 \\
Hindi & -0.001 & -0.003 & 0.002 & \text{0} & \text{0} & \text{0} & 0.711 \\
\midrule
Avg Diff & 0.001 & 0.008 & 0.001 & \text{0} & 0.002 & 0.002 & - \\
P value & \text{0.482} & \text{0.254} & \text{0.406} & \text{0.979} & \text{0.081} & - & \text{0.094} \\
\bottomrule
\end{tabular}%
}
\caption{Soft-Suff and Soft-Comp difference between mBERT and  monolingual BERT ({\small \textcolor{purple}{plum} indicates monolingual models are more faithful than multilingual models}.)
}
\label{tab:soft_bert}
\end{table}

\UseRawInputEncoding
\begin{table}[!h]
\centering
\resizebox{0.95\columnwidth}{!}{%
\begin{tabular}{@{}c|RRRRRRccccc|cc|@{}}
\hline
\multicolumn{8}{c}{\textbf{Soft-Sufficiency}} \\     
\hline
\toprule
         & $\alpha$ & $\alpha\nabla\alpha$ & $x\nabla x$ & IG     & DL      & Avg Diff & P value \\ \midrule
English & 0.07 & 0.112 & 0.006 & -0.16 & 0.035 & 0.013 & 0.699 \\
Chinese & -0.022 & 0.115 & -0.019 & 0.033 & 0.15 & 0.052 & 0.245 \\
Spanish & -0.026 & 0.08 & 0.061 & -0.154 & -0.021 & -0.012 & 0.692 \\
French & -0.045 & 0.111 & -0.094 & 0.014 & -0.006 & -0.004 & 0.893 \\
Hindi & 0.044 & -0.051 & 0.082 & \text{0} & 0.064 & 0.028 & 0.356 \\
\midrule
Avg Diff & 0.004 & 0.073 & 0.007 & -0.053 & 0.044 & 0.015 & - \\
P value & \text{0.901} & \text{0.053} & \text{0.787} & \text{0.165} & \text{0.075} & - & \text{0.3} \\
\hline
\hline
\multicolumn{8}{c}{\textbf{Soft-Comprehensiveness}} \\     
\hline
\toprule
         & $\alpha$ & $\alpha\nabla\alpha$ & $x\nabla x$ & IG     & DL      & Avg Diff & P value \\ \midrule
English & -0.006 & -0.006 & -0.008 & -0.007 & 0.001 & -0.005 & 0.105 \\
Chinese & -0.012 & -0.002 & -0.005 & -0.004 & -0.004 & -0.005 & 0.011 \\
Spanish & -0.027 & -0.007 & 0.011 & -0.064 & -0.086 & -0.035 & 0.102 \\
French & 0.003 & 0.004 & 0.009 & \text{0} & 0.005 & 0.004 & 0.19 \\
Hindi & 0.002 & \text{0} & -0.028 & -0.006 & -0.035 & -0.013 & 0.112 \\ 
\midrule
Avg Diff & -0.008 & -0.002 & -0.004 & -0.016 & -0.024 & -0.011 & - \\
P value & \text{0.143} & \text{0.346} & \text{0.564} & \text{0.203} & \text{0.181} & - & \text{0.018} \\
\bottomrule
\end{tabular}%
}
\caption{Soft-Suff and Soft-Comp difference between XLM-R and monolingual RoBERTa.)
}
\label{tab:soft_roberta}
\end{table}

\section{RoBERTa vs. BERT}\label{app:roberta_vs_bert}

\subsection{Language distribution of mBERT and XLM-R pre-training corpora}\label{app:robert_bert_language_size}
\begin{figure*}[!h] 
\centering
    \includegraphics[trim={0 0 0 0},clip, width=.89\textwidth]{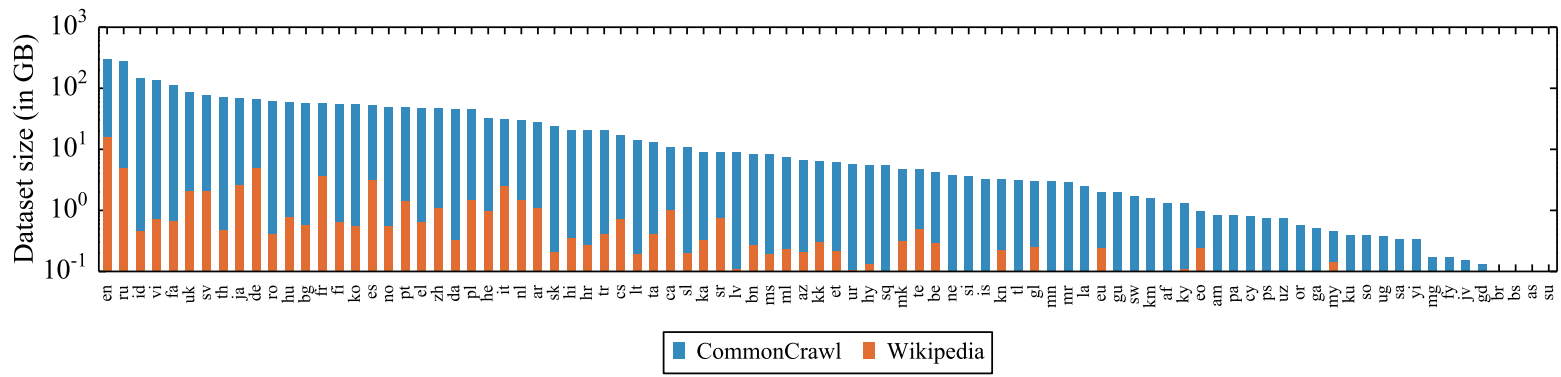}
    \caption{Amount of data in GiB (log-scale) for the 88 languages that appear in both the Wiki-100 corpus (used for mBERT) and the CC-100 (XLM-R). CC-100 increases the amount of data by several orders of magnitude, in particular for low-resource languages \cite{conneau-etal-2020-unsupervised}.}\vspace*{0pt}
    \label{fig:robert_bert_language_size}
\end{figure*}

Figure \ref{fig:robert_bert_language_size} compares the amount of data and its distribution in different languages between mBERT and XLM-R. As shown in Figure~\ref{fig:robert_bert_language_size}, XLM-R pre-training data is several orders of magnitude larger in all languages and includes a relatively higher percentage of non-English data than mBERT pre-training data.

\subsection{Full faithfulness results of XLM-R large}\label{app:full_result_robert_large}

Table \ref{tab:full_result_robert_large} presents the original Suff and Comp results of each feature attribution on each task for XLM-R large. 

\begin{table*}[!t]
\resizebox{\textwidth}{!}{%
\begin{tabular}{@{}llllllllllll@{}}
\toprule
\textbf{Dataset} & \textbf{Model} & \textbf{$\alpha$ Suff} & \textbf{$\alpha\nabla\alpha$ Suff} & \textbf{$x\nabla x$ Suff} &\textbf{ IG Suff}  & \textbf{DL Suff} & \textbf{$\alpha$ Comp} & \textbf{$\alpha\nabla\alpha$ Comp} & \textbf{$x\nabla x$ Comp} &\textbf{ IG Comp}  & \textbf{DL Comp} \\ \midrule

SST & XLM-R large & 0.9555 & 0.9547 & 1.0189 & 0.7746 & 1.0062 & 0.9437 & 0.9382 & 1.1265 & 0.6697 & 1.0576  \\
Agnews & XLM-R large & 1.1866 & 1.2698 & 0.7601 & 0.8642 & 0.9089 & 2.8766 & 2.6539 & 1.3965 & 1.3442 & 1.0955  \\
MultiRC & XLM-R large & 1.0007 & 1.0004 & 1.0007 & 0.9967 & 1.4006 & 0.8311 & 0.6314 & 0.6188 & 3.2761 & 0.6126  \\ \hline

KR & XLM-R large & 1.1857 & 1.1985 & 1.0159 & 0.9569 & 0.9741 & 1.0487 & 1.0408 & 1.0543 & 1.1403 & 1.0179 \\
ANT & XLM-R large & 1.0355 & 1.0395 & 0.9159 & 0.7393 & 1.0027 & 1.0278 & 1.0178 & 0.887 & 0.6333 & 1.0025  \\
ChnSentiCorp & XLM-R large & 0.8405 & 0.8372 & 1.044 & 0.918 & 0.9405 & 0.7424 & 0.7871 & 1.1985 & 0.9229 & 1.0699  \\ \hline

Spanish CSL & XLM-R large & 1.2667 & 1.2688 & 0.9961 & 0.9862 & 1.0137 & 1.2989 & 1.304 & 1.0417 & 1.0722 & 1.0519  \\
Spanish XNLI & XLM-R large & 0.8986 & 0.8959 & 1.0609 & 0.9614 & 0.9873 & 0.8655 & 0.8668 & 1.1609 & 1.0007 & 1.0213  \\
Spanish Paws & XLM-R large & 0.8478 & 0.8579 & 1.0342 & 0.9004 & 0.9432 & 1.1443 & 1.1448 & 1.1444 & 1.0152 & 1.0204  \\ \hline

French CSL & XLM-R large & 1.0388 & 1.0278 & 1.1031 & 1.0849 & 1.0313 & 1.0364 & 1.0361 & 1.0631 & 1.1244 & 1.0435  \\
French XNLI & XLM-R large & 1.0388 & 1.0403 & 1.079 & 0.9644 & 0.9943 & 1.0899 & 1.085 & 1.1397 & 1.0307 & 1.0227  \\
French Paws & XLM-R large & 0.8575 & 0.8583 & 1.051 & 0.9031 & 1.0132 & 1.1394 & 1.1289 & 1.2237 & 0.9642 & 1.0129  \\ \hline

Hindi BBC Nli & XLM-R large & 0.8731 & 0.8478 & 1.0379 & 1.0646 & 0.9734 & 0.7646 & 0.7796 & 1.0062 & 1.0786 & 1.0222  \\
Hindi BBC Topic & XLM-R large & 1.6458 & 1.6491 & 0.9722 & 0.8833 & 1.0009 & 1.7309 & 1.7246 & 0.9697 & 0.9469 & 1.0661  \\
Hindi XNLI & XLM-R large & 0.9875 & 0.995 & 1.0806 & 0.9227 & 0.947 & 1.0358 & 1.0309 & 1.1539 & 0.9326 & 0.9913  \\
\bottomrule
\end{tabular}%
}
\caption{Full results of faithfulness for XLM-R large. All faithfulness scores are divided by the random baseline. }
\label{tab:full_result_robert_large}
\end{table*}

\section{Impact of model size}\label{app:impact_of_model_size}

\begin{table}[]
\centering
\resizebox{0.95\columnwidth}{!}{%
\begin{tabular}{@{}c|RRRRRRccccc|cc|@{}}
\hline
\multicolumn{8}{c}{\textbf{Sufficiency}} \\     
\hline
\toprule
         & $\alpha$ & $\alpha\nabla\alpha$ & $x\nabla x$ & IG     & DL     & Avg Diff & P value \\ \midrule
English & -0.360 & -0.354 & -0.124 & -0.445 & -0.214 & -0.300 & 0.001 \\
Chinese & -0.143 & -0.133 & -0.042 & -0.220 & -0.044 & -0.116 & 0.157 \\
Spanish & -0.172 & -0.240 & -0.352 & -0.278 & -0.160 & -0.240 & 0.001 \\
French & -0.309 & -0.314 & -0.120 & -0.248 & -0.188 & -0.236 & 0.000 \\
Hindi & 0.010 & 0.012 & 0.039 & -0.239 & 0.001 & -0.035 & 0.711 \\ \midrule
Avg Diff & -0.195 & -0.206 & -0.120 & -0.286 & -0.121 & -0.186 & - \\
P value  & \text{0.057}  & \text{0.050}  & \text{0.045}  & \text{0.000}  & \text{0.035}  & -    & \text{0.000}   \\ \hline
\hline
\multicolumn{8}{c}{\textbf{Comprehensiveness}} \\     
\hline
\toprule
         & $\alpha$ & $\alpha\nabla\alpha$ & $x\nabla x$  & IG     & DL      & Avg Diff & P value \\ \midrule
English & -0.201 & -0.314 & -0.366 & 0.078 & -0.448 & -0.250 & 0.204 \\
Chinese & -0.266 & -0.254 & -0.047 & -0.303 & -0.048 & -0.183 & 0.055 \\
Spanish & -0.184 & -0.102 & -0.003 & -0.029 & -0.177 & -0.099 & 0.060 \\
French & -0.484 & -0.484 & -0.124 & -0.627 & -0.449 & -0.434 & 0.005 \\
Hindi & 0.103 & 0.091 & -0.022 & -0.364 & 0.101 & -0.018 & 0.868 \\ \midrule
Avg Diff & -0.206 & -0.212 & -0.112 & -0.249 & -0.204 & -0.197 & -   \\
P value  & \text{0.147}  & \text{0.119}  & \text{0.088}  & \text{0.169}  & \text{0.088}  & -    & \text{0.001}  \\ \bottomrule

\end{tabular}%
}
\caption{Suff and Comp difference between XLM-R Large and monolingual RoBERTa.
}
\label{tab:roberta_large}
\end{table}

\begin{table}[]
\centering
\resizebox{\columnwidth}{!}{%
\begin{tabular}{@{}l|lllll|lll@{}}
\toprule
\multicolumn{9}{c}{\textbf{Sufficiency}} \\ \midrule
 & $\alpha$ & \multicolumn{1}{c}{$\alpha\nabla\alpha$} & $x\nabla x$ & IG & DL & SST & Agnews & MultiRC \\ \midrule
BERT base (109M) & 1.279 & 1.272 & 1.005 & 1.127 & 1.044 & 1.122 & 1.061 & 1.253 \\
BERT large (340M) & 1.045 & 1.037 & 1.005 & 1.158 & 1.025 & 1.017 & 1.041 & 1.105 \\ \midrule
\multicolumn{9}{c}{\textbf{Comprehensiveness}} \\ \midrule
 & $\alpha$ & \multicolumn{1}{c}{$\alpha\nabla\alpha$} & $x\nabla x$ & IG & DL & SST & Agnews & MultiRC \\ \midrule
BERT base (109M) & 1.699 & 1.717 & 1.233 & 1.694 & 1.281 & 1.431 & 2.146 & 0.997 \\
BERT large (340M) & 1.564 & 1.581 & 1.134 & 1.731 & 1.053 & 1.270 & 1.963 & 1.005 \\ \bottomrule
\end{tabular}%
}
\caption{Suff and Comp of BERT-base and BERT-large models averaged across each FA and each task.}
\label{tab:added_bert_base_large_comparision}
\end{table}

The results indicate a lower faithfulness of the larger BERT model across FAs and tasks. Specifically, Suff and Comp scores of the monolingual English BERT-large are higher than its counterpart BERT-base (13 out of 16 comparison pairs as shown in Table~\ref{tab:added_bert_base_large_comparision}), except for cases of Suff and Comp on IG and the comprehensiveness on MultiRC (where the faithfulness of both BERT-base and large are on par with the random baseline, i.e. values close to one). This observation agrees with our assumption above that model sizes might impact faithfulness disparity. Given that our focus is on faithfulness disparity, we leave a more comprehensive study on the impact of model size on faithfulness for future work.

\section{The tokenization for different languages}
\label{app:tokenizer_subword}
\UseRawInputEncoding

All monolingual and multilingual BERT tokenizers in this paper use ``\#\#'' to indicate the second and the rest subwords of a split word, i.e. non-first subword of a split word. For example, ``sdfnsksi cklx'' will be tokenize to `sd', `\#\#fn', `\#\#sk', `\#\#si', `ck', `\#\#l', `\#\#x'.

Monolingual RoBERTa indicates a space and its following word with 'G'. Therefore, except for the first token, tokens without 'G' are subwords. XLM-R uses ``\_'' to indicate the start of a whole word.

\section{Fertility and splitting ratio}
\label{app:fertility_full_results}

\subsection{Full results}

\begin{table*}[]
\resizebox{\textwidth}{!}{%
\begin{tabular}{@{}lrrrrrrrrrrrr@{}}
\toprule
 & \multicolumn{3}{c}{\textbf{RoBERTa}} & \multicolumn{3}{c}{\textbf{BERT}} & \multicolumn{3}{c}{\textbf{RoBERTa}} & \multicolumn{3}{c}{\textbf{BERT}} \\ \midrule
Dataset & \multicolumn{1}{l}{Multi Fertility} & \multicolumn{1}{l}{Mono Fertility} & \multicolumn{1}{l}{Fertility Diff} & \multicolumn{1}{l}{Multi Fertility} & \multicolumn{1}{l}{Mono Fertility} & \multicolumn{1}{l}{Fertility Diff} & \multicolumn{1}{l}{Multi Splitting Ratio} & \multicolumn{1}{l}{Mono Splitting Ratio} & \multicolumn{1}{l}{Splitting Diff} & \multicolumn{1}{l}{Multi Splitting Ratio} & \multicolumn{1}{l}{Mono Splitting Ratio} & \multicolumn{1}{l}{Splitting Diff} \\ \midrule
SST & 1.2941 & 1.1327 & 0.1615 & 1.2229 & 1.1237 & 0.0992 & 0.2358 & 0.0893 & 0.1466 & 0.1674 & 0.0863 & 0.0811 \\
Agnews & 1.3392 & 1.1519 & 0.1873 & 1.1780 & 1.1325 & 0.0455 & 0.2724 & 0.0765 & 0.1959 & 0.0884 & 0.0504 & 0.0380 \\
MultiRC & 1.3250 & 1.0901 & 0.2350 & 1.1365 & 1.0890 & 0.0475 & 0.2734 & 0.0618 & 0.2116 & 0.0768 & 0.0397 & 0.0371 \\\hline
Spanish CSL & 1.3418 & 1.2018 & 0.1399 & 1.3796 & 1.2138 & 0.1658 & 0.2587 & 0.1596 & 0.0991 & 0.1716 & 0.0618 & 0.1098 \\
Spanish PAWS-X & 1.4706 & 1.4286 & 0.0419 & 1.3605 & 1.4034 & -0.0429 & 0.3203 & 0.2441 & 0.0762 & 0.1303 & 0.1406 & -0.0103 \\
Spanish XNLI & 1.4134 & 1.2387 & 0.1747 & 1.3679 & 1.2317 & 0.1362 & 0.3173 & 0.1819 & 0.1355 & 0.1543 & 0.0675 & 0.0868 \\\hline
French CSL & 1.4511 & 1.3134 & 0.1377 & 1.4668 & 1.3768 & 0.0900 & 0.2921 & 0.1904 & 0.1016 & 0.1553 & 0.1091 & 0.0462 \\
French PAWS-X & 1.5818 & 1.3652 & 0.2166 & 1.4257 & 1.5555 & -0.1298 & 0.3511 & 0.2195 & 0.1316 & 0.1257 & 0.1921 & -0.0664 \\
French XNLI & 1.5598 & 1.3557 & 0.2041 & 1.4912 & 1.4353 & 0.0558 & 0.3307 & 0.2233 & 0.1074 & 0.1358 & 0.1011 & 0.0347 \\ \bottomrule
\end{tabular}%
}
\caption{Fertility and splitting ratio of multilingual and monolingual RoBERTa and BERT on various tasks.}
\label{tab:fertility_full_results}
\end{table*}

Table \ref{tab:fertility_full_results} includes the full results of fertility and splitting ratio for each model. These are used for calculating the average values in Table \ref{tab:tokenizer_diff}.

\subsection{Results on Hindi}

Table~\ref{tab:hindi_split_ratio} shows the splitting ratio and fertility rate for Hindi, where lower scores indicate that the tokenizer is less aggressive and better suited to the language. Hindi does not show a consistent pattern between multi- and monolingual settings in terms of tokenization aggressiveness. For example, the splitting ratio of XLM-R (avg. 0.331) is less aggressive than Hindi RoBERTa (avg. 0.869), while mBERT (0.394) is slightly more aggressive than Hindi BERT (0.267). This observation on Hindi is different from the three languages in Table~\ref{tab:tokenizer_diff} (multilingual tokenizers are more aggressive), which indicates a potential opportunity for future research, e.g. exploring whether or how the aggressiveness of tokenization impacts faithfulness for different languages.

\begin{table*}[!ht]
\resizebox{\textwidth}{!}{%
\begin{tabular}{@{}l|ccc|ccc|ccc|ccc@{}}
\toprule
                & \multicolumn{3}{c|}{RoBERTa}                      & \multicolumn{3}{c|}{BERT}                         & \multicolumn{3}{c|}{RoBERTa}                                  & \multicolumn{3}{c}{BERT}                                      \\ \midrule
Dataset         & Multi Fertility & Mono Fertility & Fertility Diff & Multi Fertility & Mono Fertility & Fertility Diff & Multi Splitting Ratio & Mono Splitting Ratio & Splitting Diff & Multi Splitting Ratio & Mono Splitting Ratio & Splitting Diff \\
Hindi BBC Nli   & 1.677           & 3.837          & -2.160         & 2.129           & 1.570          & 0.560          & 0.400                 & 0.828                & -0.428         & 0.488                 & 0.290                & 0.198          \\
Hindi BBC Topic & 1.467           & 3.560          & -2.093         & 1.844           & 1.572          & 0.272          & 0.303                 & 0.813                & -0.510         & 0.359                 & 0.261                & 0.098          \\
Hindi XNLI      & 1.429           & 3.621          & -2.192         & 1.749           & 1.506          & 0.243          & 0.291                 & 0.968                & -0.677         & 0.335                 & 0.250                & 0.085          \\
Avg             & 1.524           & 3.673          & -2.148         & 1.908           & 1.549          & 0.358          & 0.331                 & 0.869                & -0.538         & 0.394                 & 0.267                & 0.127          \\ \bottomrule
\end{tabular}%
}
\caption{ The splitting ratio and fertility rate for Hindi on the three Hindi datasets.}
\label{tab:hindi_split_ratio}
\end{table*}

\section{Disparity in tokenization aggressiveness}\label{app:SentencePiece}

\begin{figure}[!h] 
\centering
    \includegraphics[trim={0 0 0 2cm},clip, width=.9\columnwidth]{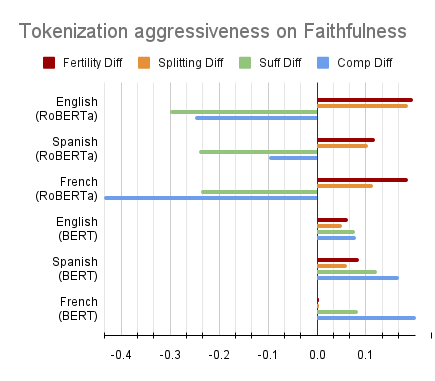}
    \caption{The impact of tokenization aggressiveness ("Fertility Diff" and "Splitting Diff") on faithfulness disparity (``Suff Diff'' and ``Comp Diff'').}\vspace*{0pt}
    \label{fig:SentencePiece}
\end{figure}

Figure \ref{fig:SentencePiece} shows the difference between multi- and monolingual models in terms of tokenization aggressiveness and faithfulness. Both are calculated as the score of the multilingual model minus the corresponding score of the monolingual counterpart model. We observe that multilingual models consistently tokenize more aggressively than their monolingual counterparts. When the fertility of the multilingual model is higher than its monolingual counterpart (i.e. by more than 0.1), the multilingual model gains lower faithfulness.

\subsection{Implementation of WECHSEL and FOCUS}\label{app:french_focus_implementation}

WECHSEL first copies all non-embedding parameters from the English RoBERTa, and replaces the tokenizer with the tokenizer of French RoBERTa. In this paper, we use the WECHSEL model in French published by \citet{minixhofer-etal-2022-wechsel}.\footnote{\url{https://github.com/CPJKU/wechsel}}.

FOCUS extends the embedding matrix of XLM-R with non-overlapping tokens from the French RoBERTa tokenizer. These new tokens are represented as the weighted mean of overlapping tokens' embeddings. The overlapping tokens between both tokenizers are the anchor points to find the similar tokens for calculating the weighted mean. 
For FOCUS, we use FastText embeddings in French to obtain the static token embeddings and find similar tokens following \citep{dobler-demelo-2023-focus}. We use the default hyperparameter settings in FOCUS.\footnote{\url{https://github.com/konstantinjdobler/focus}}

\section{Full Results of Faithfulness}
\label{app:full_results}

Table \ref{tab:full_results} shows the Suff and Comp of each feature attribution on each dataset. Table \ref{tab:full_soft_results}  shows the Soft-Suff and Soft-Comp of each feature attribution on each dataset. All faithfulness scores are presented as ratios after being divided by the random baseline. The predictive results, F1 and accuracy, are the average over three runs. The best model from the three runs is taken to extract and evaluate the rationales with each feature attribution method separately. 
    
\begin{table*}[!t]
\resizebox{\textwidth}{!}{%
\begin{tabular}{@{}llllllllllllll@{}}
\toprule
\textbf{Dataset} & \textbf{Model} & \textbf{$\alpha$ Suff} & \textbf{$\alpha\nabla\alpha$ Suff} & \textbf{$x\nabla x$} \textbf{Suff} & \textbf{IG Suff}  & \textbf{DL Suff} & \textbf{$\alpha$ Comp} & \textbf{$\alpha\nabla\alpha$ Comp} & \textbf{$x\nabla x$ Comp} & \textbf{IG Comp}  & \textbf{DL Comp} & \textbf{F1} & \textbf{Accuracy} \\ \midrule
SST & mBERT & 1.2063 & 1.205 & 0.9991 & 1.3995 & 1.2594 & 1.2576 & 1.2643 & 1.0433 & 1.4835 & 1.3135 & 0.8627 & 0.8627 \\
SST & XLM-R & 1.0914 & 1.0976 & 1.0329 & 1.1125 & 1.0558 & 0.9242 & 0.9244 & 0.9537 & 1.0787 & 0.9878 & 0.8718 & 0.8719 \\
SST & BERT & 1.174 & 1.1771 & 1.0207 & 1.1636 & 1.0726 & 1.5571 & 1.5597 & 1.1582 & 1.6837 & 1.1955 & 0.9156 & 0.9156 \\
SST & RoBERTa & 1.2623 & 1.2693 & 1.3215 & 1.4922 & 1.1866 & 1.6021 & 1.6144 & 1.2723 & 1.438 & 1.3409 & 0.8893 & 0.8898 \\

Agnews & mBERT & 1.7087 & 1.712 & 0.9817 & 1.4523 & 1.0573 & 3.2063 & 3.203 & 1.8811 & 2.8304 & 1.5659 & 0.9303 & 0.9304 \\
Agnews & XLM-R & 2.0947 & 2.105 & 0.9287 & 1.4987 & 0.8806 & 2.0106 & 2.0107 & 1.2924 & 1.9369 & 1.1211 & 0.9261 & 0.9264 \\
Agnews & BERT & 1.1553 & 1.1266 & 0.9105 & 1.0425 & 1.0719 & 2.5436 & 2.5968 & 1.5426 & 2.4037 & 1.6445 & 0.9357 & 0.9357 \\
Agnews & RoBERTa & 1.3137 & 1.3242 & 0.8989 & 1.452 & 1.4351 & 2.1323 & 2.1408 & 1.66 & 1.9998 & 1.0854 & 0.9347 & 0.9346 \\

MultiRC & mBERT & 1.1821 & 1.177 & 0.9611 & 1.0904 & 0.9612 & 1.0 & 1.0011 & 1.0004 & 1.0031 & 1.0065 & 0.7081 & 0.7186 \\
MultiRC & XLM-R & 0.7907 & 0.829 & 0.9001 & 0.9677 & 1.0648 & 0.9247 & 0.9204 & 1.0124 & 1.0424 & 1.0109 & 0.718 & 0.7245 \\
MultiRC & BERT & 1.5089 & 1.512 & 1.0829 & 1.1752 & 0.9888 & 0.9959 & 0.9948 & 0.9978 & 0.9942 & 1.0022 & 0.6815 & 0.6896 \\
MultiRC & RoBERTa & 1.648 & 1.6946 & 0.9313 & 1.0268 & 1.3368 & 1.5195 & 1.4091 & 1.3068 & 1.6189 & 1.6841 & 0.7295 & 0.7317 \\

KR & mBERT & 1.1229 & 1.0541 & 1.1878 & 1.3514 & 1.1128 & 1.0077 & 1.0082 & 0.9979 & 0.9989 & 0.9966 & 0.842 & 0.8424 \\
KR & XLM-R & 1.4342 & 1.4154 & 0.8885 & 1.0773 & 0.938 & 0.9022 & 0.9014 & 1.0259 & 1.0089 & 1.0307 & 0.8401 & 0.8403 \\
KR & BERT (zh) & 1.239 & 1.2241 & 1.0296 & 1.0242 & 0.9226 & 1.0105 & 1.0157 & 0.996 & 0.9907 & 1.0165 & 0.8399 & 0.84 \\
KR & RoBERTa (zh) & 0.8657 & 0.8376 & 1.0082 & 0.9963 & 0.9782 & 0.9912 & 0.9932 & 0.9882 & 0.9901 & 0.9989 & 0.8443 & 0.8446 \\

ANT & mBERT & 1.0425 & 1.0471 & 0.9258 & 0.9767 & 0.8555 & 1.049 & 1.0455 & 1.0228 & 1.0208 & 1.0915 & 0.6282 & 0.703 \\
ANT & XLM-R & 1.0033 & 0.991 & 0.948 & 1.0205 & 1.0631 & 0.953 & 0.9601 & 0.9287 & 0.9879 & 1.0229 & 0.6588 & 0.7139 \\
ANT & BERT (zh) & 1.2248 & 1.2319 & 0.9675 & 1.0107 & 0.9884 & 1.0216 & 1.0212 & 1.0032 & 1.0105 & 1.0051 & 0.6738 & 0.7237 \\
ANT & RoBERTa (zh) & 1.0773 & 1.0945 & 1.0446 & 1.1371 & 1.1157 & 1.0063 & 1.0033 & 1.0057 & 1.0261 & 1.0252 & 0.5241 & 0.6601 \\

ChnSentiCorp & mBERT & 1.4906 & 1.4942 & 1.0566 & 1.325 & 1.0146 & 2.1555 & 2.1608 & 1.324 & 2.0856 & 1.0983 & 0.9119 & 0.9119 \\
ChnSentiCorp & XLM-R & 1.2483 & 1.2368 & 1.0077 & 1.055 & 0.9944 & 1.0723 & 1.0738 & 0.9931 & 1.1389 & 0.9942 & 0.9217 & 0.9217 \\
ChnSentiCorp & BERT (zh) & 1.2466 & 1.2516 & 1.0455 & 1.09 & 1.0243 & 1.548 & 1.5388 & 1.2609 & 1.5762 & 1.1181 & 0.9355 & 0.9356 \\
ChnSentiCorp & RoBERTa (zh) & 1.5482 & 1.5435 & 1.0476 & 1.1406 & 0.9543 & 1.6196 & 1.6116 & 1.2854 & 1.5884 & 1.2097 & 0.9428 & 0.9428 \\

Spanish CSL & mBERT & 1.5244 & 1.5274 & 1.0999 & 1.6256 & 1.1076 & 1.898 & 1.8972 & 1.2135 & 1.9047 & 1.2905 & 0.886 & 0.8862 \\
Spanish CSL & XLM-R & 1.1065 & 1.0896 & 0.9543 & 1.1994 & 1.0514 & 0.986 & 0.9887 & 0.9715 & 1.1801 & 0.9913 & 0.878 & 0.8782 \\
Spanish CSL & BERT (es) & 0.9975 & 0.976 & 0.9957 & 1.1277 & 1.0645 & 1.0698 & 1.0788 & 1.0955 & 1.4271 & 1.0004 & 0.9062 & 0.9063 \\
Spanish CSL & RoBERTa (es) & 1.2901 & 1.4932 & 1.5522 & 1.5633 & 1.5125 & 1.5761 & 1.3826 & 1.3995 & 1.0484 & 1.5366 & 0.8914 & 0.8917 \\

Spanish XNLI & mBERT & 1.0031 & 1.0043 & 1.0258 & 1.0382 & 1.0331 & 1.0165 & 1.0164 & 0.9964 & 1.0028 & 0.9872 & 0.7877 & 0.7875 \\
Spanish XNLI & XLM-R & 1.0314 & 1.0457 & 1.0887 & 1.0738 & 1.0521 & 1.0485 & 1.0479 & 1.0285 & 1.0469 & 0.9918 & 0.7958 & 0.7956 \\
Spanish XNLI & BERT (es) & 1.0791 & 1.0922 & 1.037 & 1.0228 & 1.0331 & 1.0327 & 1.03 & 0.9938 & 1.0017 & 0.9721 & 0.7847 & 0.7842 \\
Spanish XNLI & RoBERTa (es) & 1.3083 & 1.3127 & 1.5799 & 1.1294 & 0.9508 & 1.102 & 1.1 & 1.0525 & 1.0146 & 1.0096 & 0.7958 & 0.7956 \\

Spanish Paws & mBERT & 1.1325 & 1.1348 & 0.9959 & 0.9616 & 0.9826 & 0.994 & 0.9952 & 0.9968 & 0.9999 & 1.0062 & 0.8811 & 0.8823 \\
Spanish Paws & XLM-R & 1.1797 & 1.1944 & 1.0948 & 1.0857 & 0.9884 & 1.2369 & 1.2376 & 1.0415 & 1.0452 & 0.987 & 0.8703 & 0.872 \\
Spanish Paws & BERT (es) & 0.9825 & 0.9919 & 1.0713 & 0.9047 & 0.9792 & 1.0016 & 0.997 & 0.9985 & 0.9988 & 0.9965 & 0.8555 & 0.8565 \\
Spanish Paws & RoBERTa (es) & 0.9294 & 0.9379 & 1.0151 & 0.9883 & 0.9621 & 1.1832 & 1.1391 & 0.9047 & 1.1132 & 1.0781 & 0.8823 & 0.883 \\

French CSL & mBERT & 1.4165 & 1.413 & 0.9956 & 1.4875 & 1.1035 & 2.1526 & 2.1624 & 1.1415 & 2.0983 & 1.3063 & 0.8772 & 0.8773 \\
French CSL & XLM-R & 1.1488 & 1.16 & 0.9952 & 1.0022 & 1.0042 & 0.9769 & 0.9721 & 1.0087 & 1.1822 & 0.9862 & 0.8863 & 0.8865 \\
French CSL & BERT (fr) & 1.0753 & 1.0857 & 0.9524 & 1.2311 & 0.8271 & 1.2186 & 1.211 & 0.9881 & 1.4274 & 0.852 & 0.8824 & 0.8825 \\
French CSL & RoBERTa (fr) & 1.3471 & 1.3482 & 1.1526 & 1.4631 & 1.4639 & 2.0347 & 2.0311 & 1.4313 & 2.5163 & 2.3467 & 0.8663 & 0.8668 \\

French XNLI & mBERT & 1.0997 & 1.0732 & 1.0201 & 1.1127 & 1.0719 & 1.0147 & 1.0175 & 0.9985 & 1.0194 & 1.0179 & 0.7748 & 0.7746 \\
French XNLI & XLM-R & 1.0058 & 0.9517 & 1.1456 & 1.0234 & 1.0441 & 1.0544 & 1.0577 & 1.0324 & 1.027 & 0.9889 & 0.789 & 0.7885 \\
French XNLI & BERT (fr) & 0.9795 & 0.9862 & 1.0337 & 1.0762 & 1.0819 & 1.0503 & 1.0484 & 1.0077 & 1.0389 & 0.9974 & 0.7643 & 0.7638 \\
French XNLI & RoBERTa (fr) & 1.5508 & 1.5543 & 1.4092 & 1.183 & 1.1098 & 1.527 & 1.5246 & 1.2518 & 1.0796 & 0.9975 & 0.7326 & 0.7323 \\

French Paws & mBERT & 1.1789 & 1.1849 & 0.9801 & 0.9469 & 0.8695 & 0.9808 & 0.9798 & 0.9963 & 0.998 & 1.0062 & 0.8781 & 0.8788 \\
French Paws & XLM-R & 1.087 & 1.1021 & 1.0529 & 1.0192 & 0.9929 & 1.2295 & 1.2255 & 1.0622 & 0.997 & 1.0263 & 0.8774 & 0.8778 \\
French Paws & BERT (fr) & 1.0875 & 1.0796 & 1.0948 & 1.0518 & 1.0596 & 0.9987 & 1.0022 & 1.0028 & 0.9994 & 1.0144 & 0.8274 & 0.8297 \\
French Paws & RoBERTa (fr) & 0.9629 & 0.9655 & 1.0318 & 1.0507 & 1.0304 & 1.1575 & 1.1452 & 1.1168 & 1.4052 & 1.0831 & 0.7729 & 0.8678 \\

Hindi BBC Nli & mBERT & 1.1255 & 1.1278 & 1.1362 & 1.175 & 1.0102 & 1.0044 & 1.0039 & 1.003 & 0.998 & 1.005 & 0.7862 & 0.7864 \\
Hindi BBC Nli & XLM-R & 1.1809 & 1.1789 & 1.0289 & 1.0762 & 1.0578 & 1.18 & 1.19 & 1.0317 & 1.0842 & 1.0125 & 0.7887 & 0.7888 \\
Hindi BBC Nli & BERT (hi) & 0.9799 & 0.9779 & 1.0385 & 1.0574 & 1.0385 & 1.0122 & 1.016 & 0.9989 & 1.0046 & 1.0045 & 0.8124 & 0.8128 \\
Hindi BBC Nli & RoBERTa (hi) & 1.0349 & 1.0225 & 0.9337 & 0.9863 & 0.9436 & 0.6561 & 0.6876 & 1.1159 & 1.0714 & 0.9546 & 0.7953 & 0.8094 \\

Hindi BBC Topic & mBERT & 1.4883 & 1.4913 & 1.2533 & 1.3573 & 0.984 & 1.4896 & 1.4907 & 1.2431 & 1.2887 & 1.0935 & 0.5123 & 0.5918 \\
Hindi BBC Topic & XLM-R & 1.1243 & 1.1513 & 1.0942 & 1.2419 & 1.1083 & 1.1409 & 1.1413 & 1.0351 & 1.1042 & 1.0049 & 0.5606 & 0.6425 \\
Hindi BBC Topic & BERT (hi) & 1.8746 & 1.8729 & 1.498 & 0.9446 & 1.0336 & 2.1692 & 2.1703 & 1.6329 & 0.8877 & 0.8943 & 0.617 & 0.6753 \\
Hindi BBC Topic & RoBERTa (hi) & 0.9569 & 0.9527 & 0.9921 & 1.2189 & 0.9464 & 0.9823 & 0.9841 & 1.04 & 1.4999 & 0.9481 & 0.5268 & 0.6395 \\

Hindi XNLI & mBERT & 1.1363 & 1.1501 & 1.2088 & 1.3084 & 1.071 & 1.0187 & 1.0159 & 1.0147 & 1.0359 & 0.9775 & 0.6754 & 0.676 \\
Hindi XNLI & XLM-R & 1.0099 & 0.9844 & 0.985 & 1.0652 & 0.9954 & 1.1142 & 1.1161 & 1.0214 & 1.0578 & 1.0056 & 0.7235 & 0.7237 \\
Hindi XNLI & BERT (hi) & 1.0199 & 1.0234 & 1.0304 & 1.0419 & 1.0032 & 1.0266 & 1.0248 & 1.0126 & 1.0165 & 1.0048 & 0.6607 & 0.6607 \\
Hindi XNLI & RoBERTa (hi) & 1.4853 & 1.4795 & 1.0466 & 1.3833 & 1.0287 & 1.5834 & 1.589 & 1.0399 & 1.4781 & 0.8741 & 0.6316 & 0.6314 \\ \bottomrule
\end{tabular}%
}
\caption{Full results of Suff and Comp, and predictive performance. All faithfulness scores are divided by the random baseline.}
\label{tab:full_results}
\end{table*}

\begin{table*}[!t]
\resizebox{.98\textwidth}{!}{%
\begin{tabular}{@{}llllllllllllll@{}}
\toprule
\textbf{Dataset} & \textbf{Model} & \textbf{$\alpha$ Suff} & \textbf{$\alpha\nabla\alpha$ Suff} & \textbf{$x\nabla x$} \textbf{Suff} & \textbf{IG Suff}  & \textbf{DL Suff} & \textbf{$\alpha$ Comp} & \textbf{$\alpha\nabla\alpha$ Comp} & \textbf{$x\nabla x$ Comp} & \textbf{IG Comp}  & \textbf{DL Comp} \\ \midrule
SST & BERT & 1.1174 & 1.1115 & 1.0811 & 1.3015 & 1.1817 & 1.2007 & 1.2005 & 1.2002 & 1.1995 & 1.1998 \\
SST & mBERT & 1.2650 & 1.3227 & 1.0719 & 1.1442 & 1.1330 & 1.2095 & 1.2131 & 1.2053 & 1.1965 & 1.2158 \\
SST & RoBERTa & 1.1197 & 1.0435 & 1.1384 & 1.2683 & 1.1531 & 1.1977 & 1.1985 & 1.2008 & 1.1992 & 1.1970 \\
SST & XLM-R & 1.2423 & 1.1081 & 1.1750 & 1.1196 & 1.2824 & 1.2054 & 1.2038 & 1.2006 & 1.2022 & 1.2069 \\
Agnews & BERT & 1.2185 & 1.2676 & 1.0728 & 1.0625 & 1.0420 & 1.2064 & 1.2095 & 1.2079 & 1.2023 & 1.2045 \\
Agnews & mBERT & 1.1117 & 1.3246 & 1.2289 & 1.0848 & 1.0826 & 1.2014 & 1.2007 & 1.2003 & 1.2000 & 1.1996 \\
Agnews & RoBERTa & 1.0688 & 1.0713 & 1.3344 & 1.3121 & 1.2398 & 1.2016 & 1.1992 & 1.2008 & 1.1983 & 1.2024 \\
Agnews & XLM-R & 1.2147 & 1.3088 & 1.3372 & 1.0413 & 1.1902 & 1.1984 & 1.1972 & 1.1978 & 1.1996 & 1.2001 \\
MultiRC & BERT & 1.2937 & 1.2236 & 1.2478 & 1.0459 & 1.3163 & 1.1974 & 1.1963 & 1.1986 & 1.1952 & 1.2012 \\
MultiRC & mBERT & 1.1990 & 1.0981 & 1.1447 & 1.2792 & 1.2093 & 1.2023 & 1.2028 & 1.2003 & 1.2017 & 1.2033 \\
MultiRC & RoBERTa & 1.2802 & 1.1747 & 1.3438 & 1.2485 & 1.3111 & 1.2059 & 1.1938 & 1.2087 & 1.2030 & 1.1969 \\
MultiRC & XLM-R & 1.2223 & 1.2089 & 1.3215 & 1.1894 & 1.3373 & 1.1825 & 1.1719 & 1.1878 & 1.1771 & 1.1932 \\
ChnSentiCorp & BERT(zh) & 1.1092 & 1.0514 & 1.1237 & 0.9149 & 0.8834 & 1.0038 & 0.9987 & 1.0025 & 0.9975 & 1.0013 \\
ChnSentiCorp & mBERT & 1.0982 & 1.1434 & 1.1168 & 0.9094 & 0.9811 & 0.9983 & 0.9997 & 0.9990 & 0.9977 & 1.0003 \\
ChnSentiCorp & RoBERTa(zh) & 1.1400 & 0.8794 & 0.9780 & 1.1358 & 0.9518 & 1.0042 & 1.0052 & 1.0011 & 1.0032 & 1.0022 \\
ChnSentiCorp & XLM-R & 0.8697 & 1.1414 & 1.0343 & 1.1033 & 1.0253 & 0.9964 & 0.9999 & 0.9981 & 0.9973 & 1.0007 \\
KR & BERT(zh) & 0.8564 & 0.9301 & 1.0576 & 0.9246 & 1.0910 & 1.0001 & 1.0002 & 0.9999 & 0.9998 & 1.0003 \\
KR & mBERT & 0.9053 & 0.9865 & 1.1194 & 0.9745 & 0.9071 & 0.9990 & 1.0001 & 0.9997 & 1.0006 & 0.9993 \\
KR & RoBERTa(zh) & 0.9594 & 0.9000 & 1.0735 & 1.1351 & 0.9105 & 0.9991 & 1.0013 & 0.9996 & 1.0009 & 1.0004 \\
KR & XLM-R & 0.9091 & 1.1106 & 0.9407 & 1.0584 & 1.1169 & 0.9911 & 0.9938 & 1.0038 & 1.0016 & 0.9964 \\
ANT & BERT(zh) & 0.8968 & 1.0331 & 1.1381 & 1.0937 & 1.0724 & 1.0001 & 0.9014 & 1.0001 & 1.0213 & 1.0002 \\
ANT & mBERT & 0.9319 & 0.9484 & 0.8653 & 0.9424 & 1.0193 & 1.0048 & 1.0034 & 1.0071 & 1.0022 & 1.0059 \\
ANT & RoBERTa(zh) & 0.8758 & 1.0080 & 1.0925 & 0.9138 & 0.8952 & 1.0154 & 1.0031 & 1.0062 & 1.0125 & 1.0093 \\
ANT & XLM-R & 1.1315 & 0.8814 & 1.1122 & 1.1218 & 1.0655 & 0.9949 & 1.0105 & 0.9905 & 1.0069 & 1.0030 \\
Hindi XNLI & BERT(hi) & 0.9956 & 0.9982 & 0.9964 & 0.9973 & 0.9991 & 1.0038 & 1.0017 & 1.0030 & 1.0024 & 1.0009 \\
Hindi XNLI & mBERT & 0.9927 & 1.0037 & 0.9899 & 0.9955 & 0.9983 & 1.0039 & 1.0000 & 1.0049 & 1.0030 & 1.0020 \\
Hindi XNLI & RoBERTa(hi) & 0.9973 & 0.9991 & 1.0009 & 0.9982 & 0.9964 & 1.0051 & 1.0017 & 0.9984 & 1.0034 & 1.0069 \\
Hindi XNLI & XLM-R & 1.0002 & 0.9985 & 0.9996 & 1.0007 & 0.9979 & 0.9973 & 0.9999 & 0.9981 & 0.9963 & 1.0008 \\
Hindi BBC Nli & BERT(hi) & 0.8755 & 1.1450 & 1.1292 & 0.9658 & 0.9224 & 1.0033 & 1.0065 & 0.9986 & 1.0015 & 1.0050 \\
Hindi BBC Nli & mBERT & 0.9492 & 0.8837 & 1.1009 & 1.0972 & 1.0853 & 1.0001 & 0.9993 & 1.0018 & 1.0024 & 1.0030 \\
Hindi BBC Nli & RoBERTa(hi) & 0.8649 & 1.1123 & 0.9938 & 1.1217 & 0.9366 & 0.9789 & 0.9914 & 1.0917 & 1.0068 & 1.0853 \\
Hindi BBC Nli & XLM-R & 1.0760 & 0.9141 & 1.1469 & 1.0084 & 1.1297 & 0.9916 & 0.9960 & 1.0112 & 0.9975 & 0.9911 \\
Hindi BBC Topic & BERT(hi) & 0.9914 & 0.8890 & 0.9382 & 0.8912 & 1.0277 & 0.9976 & 0.9992 & 0.9969 & 1.0008 & 0.9984 \\
Hindi BBC Topic & mBERT & 1.0715 & 0.8891 & 1.0780 & 0.9357 & 1.1289 & 0.9984 & 0.9977 & 0.9991 & 1.0004 & 0.9998 \\
Hindi BBC Topic & RoBERTa(hi) & 1.0792 & 0.9567 & 1.0483 & 1.0080 & 1.0797 & 0.9978 & 1.0011 & 0.9989 & 1.0032 & 1.0022 \\
Hindi BBC Topic & XLM-R & 0.9972 & 1.0010 & 1.1432 & 1.1174 & 1.0763 & 0.9999 & 0.9981 & 0.9961 & 1.0007 & 0.9972 \\
French CSL & BERT(fr) & 0.9700 & 1.1767 & 1.1052 & 1.2127 & 1.1523 & 1.1056 & 1.0937 & 1.1078 & 1.0968 & 1.1029 \\
French CSL & mBERT & 1.1124 & 1.0003 & 1.0063 & 1.1548 & 1.1742 & 1.1014 & 1.1023 & 1.1019 & 1.1006 & 1.1028 \\
French CSL & RoBERTa(fr) & 1.0482 & 0.9482 & 1.1602 & 1.1290 & 1.0932 & 1.1006 & 1.0994 & 1.0987 & 1.0975 & 1.0981 \\
French CSL & XLM-R & 0.9456 & 1.1939 & 0.9480 & 1.0073 & 1.1293 & 1.1014 & 1.0941 & 1.1039 & 1.1064 & 1.0966 \\
French Paws & BERT(fr) & 1.0163 & 1.0772 & 1.0279 & 1.0206 & 1.2284 & 1.0972 & 1.1007 & 1.0986 & 1.0979 & 1.0993 \\
French Paws & mBERT & 1.0518 & 1.1542 & 1.1038 & 1.2155 & 1.0163 & 1.1018 & 1.1006 & 1.1022 & 1.1026 & 1.1014 \\
French Paws & RoBERTa(fr) & 1.1262 & 1.0503 & 1.0594 & 1.0303 & 1.1667 & 1.0894 & 1.0819 & 1.0678 & 1.1084 & 1.0764 \\
French Paws & XLM-R & 1.0932 & 1.1402 & 0.9883 & 1.1941 & 1.1115 & 1.1021 & 1.0968 & 1.0991 & 1.1000 & 1.0975 \\
French XNLI & BERT(fr) & 1.0956 & 1.0985 & 1.1030 & 1.0971 & 1.1015 & 1.1026 & 1.1009 & 1.0982 & 1.1017 & 1.0991 \\
French XNLI & mBERT & 1.0972 & 1.0822 & 1.0934 & 1.0897 & 1.0859 & 1.1019 & 1.1056 & 1.1028 & 1.1038 & 1.1047 \\
French XNLI & RoBERTa(fr) & 1.0994 & 1.1012 & 1.0981 & 1.1006 & 1.0987 & 1.1012 & 1.0975 & 1.1035 & 1.0988 & 1.1024 \\
French XNLI & XLM-R & 1.1003 & 1.0986 & 1.1007 & 1.0994 & 1.0999 & 1.0959 & 1.1000 & 1.0949 & 1.0980 & 1.0970 \\
Spanish CSL & BERT(es) & 1.2813 & 0.9915 & 1.2320 & 1.1154 & 1.1478 & 1.1456 & 1.1568 & 1.1545 & 1.1522 & 1.1478 \\
Spanish CSL & mBERT & 1.0206 & 1.1762 & 1.1969 & 1.0003 & 0.9930 & 1.1537 & 1.1550 & 1.1524 & 1.1564 & 1.1576 \\
Spanish CSL & RoBERTa(es) & 0.9958 & 1.0382 & 1.0512 & 1.2614 & 1.2268 & 1.1541 & 1.1562 & 1.1602 & 1.1520 & 1.1583 \\
Spanish CSL & XLM-R & 1.0035 & 1.2053 & 1.2477 & 1.0237 & 1.2107 & 1.1447 & 1.1465 & 1.1483 & 1.1474 & 1.1456 \\
Spanish Paws & BERT(es) & 1.2447 & 1.2738 & 1.0227 & 0.9928 & 1.1452 & 1.1539 & 1.1487 & 1.1526 & 1.1513 & 1.1474 \\
Spanish Paws & mBERT & 1.2031 & 1.1840 & 1.2014 & 1.0241 & 1.2006 & 1.1504 & 1.1516 & 1.1522 & 1.1498 & 1.1492 \\
Spanish Paws & RoBERTa(es) & 1.1039 & 1.0871 & 1.1247 & 1.2679 & 1.2201 & 1.2160 & 1.1559 & 1.0993 & 1.3304 & 1.3864 \\
Spanish Paws & XLM-R & 1.0206 & 1.1617 & 1.1097 & 1.0455 & 1.1732 & 1.1466 & 1.1431 & 1.1478 & 1.1458 & 1.1448 \\
Spanish XNLI & BERT(es) & 1.1556 & 1.1593 & 1.1575 & 1.1537 & 1.1519 & 1.1479 & 1.1465 & 1.1472 & 1.1486 & 1.1493 \\
Spanish XNLI & mBERT & 1.1435 & 1.1459 & 1.1485 & 1.1534 & 1.1558 & 1.1524 & 1.1519 & 1.1515 & 1.1505 & 1.1501 \\
Spanish XNLI & RoBERTa(es) & 1.1491 & 1.1497 & 1.1484 & 1.1487 & 1.1494 & 1.1534 & 1.1511 & 1.1556 & 1.1545 & 1.1523 \\
Spanish XNLI & XLM-R & 1.1475 & 1.1478 & 1.1484 & 1.1481 & 1.1487 & 1.1526 & 1.1519 & 1.1505 & 1.1512 & 1.1497 \\ \hline
\end{tabular}%
}
\caption{Full results of Soft-Suff and Soft-Comp. All faithfulness results are divided by the random baseline.}
\label{tab:full_soft_results}
\end{table*}

\end{document}